\documentclass[11pt]{article}

\PassOptionsToPackage{table}{xcolor}
\usepackage[preprint]{acl}

\usepackage{times}
\usepackage{latexsym}

\usepackage[T1]{fontenc}

\usepackage[utf8]{inputenc}

\usepackage{microtype}

\usepackage{inconsolata}

\usepackage{graphicx}
\usepackage{amsmath,amssymb,bm}

\usepackage{booktabs,multirow,makecell,array,tabularx,wrapfig}

\usepackage{pifont}
\usepackage[breakable]{tcolorbox}
\usepackage{listings}

\definecolor{mygreen}{RGB}{0,180,0}
\definecolor{myred}{RGB}{220,0,0}
\newcommand{\cmark}{\textcolor{mygreen}{\ding{51}}}
\newcommand{\xmark}{\textcolor{myred}{\ding{55}}}
\newcolumntype{C}{>{\centering\arraybackslash}X}

\newcommand{\evaluationNum}{100}
\newcommand{\questionNum}{1922}
\newcommand{\videoNum}{258}
\newcommand{\OmniscientLeakage}{Temporal Information Leakage}
\newcommand{\Perspective}{Recursive Perspective Conflation}
\newcommand{\Faulty}{Unwarranted Affective Deduction}
\newcommand{\Others}{Others}
\newcommand{\benchmarkName}{AffToM}
\newcommand{\methodName}{AffToM-R1}
\newcommand{\grpoName}{PAFRO}

\title{AffToM: Benchmarking Recursive Affective Theory of Mind in Dynamic Multi-Party Interactions}

\author{
  Jinpeng Hu\textsuperscript{1},
  Yiduo Jin\textsuperscript{1},
  Erqiang Wang\textsuperscript{1},
  Jia Li\textsuperscript{1},
  Jingjing Wu\textsuperscript{1},
  Peipei Song\textsuperscript{2} \\
  \normalfont\textsuperscript{1}Hefei University of Technology \\
  \normalfont\textsuperscript{2}University of Science and Technology of China
}

\begin{document}
\maketitle

\begin{abstract}
Understanding emotion in realistic social interactions requires more than recognizing facial expressions or assigning affective labels.
The same event can elicit different emotions depending on what each participant knows, what they believe others know or feel, and how these nested mental states evolve over time.
Existing benchmarks largely study higher-order epistemic Theory of Mind (ToM) and multimodal emotion reasoning in isolation, leaving recursive affective reasoning under asymmetric and dynamically changing information underexplored.
Therefore, we introduce {\benchmarkName}, a benchmark for affective ToM with 1,922 test pairs and 19,759 training pairs.
\benchmarkName{} evaluates Individual Affective Reasoning (IAR), which infers emotions from events and socio-cognitive context, and Multi-party Affective Reasoning (MPAR), which recursively tracks participants' nested beliefs and emotions as new evidence emerges.
We develop an event-driven multi-agent generation pipeline grounded in dynamic character interaction graphs that explicitly capture event sequences, interpersonal relations, and mental-state updates.
Furthermore, we propose \methodName{}, a two-stage post-training framework combining supervised fine-tuning with perspective-aware rubric-guided reinforcement learning.
Experiments show that MPAR remains challenging, with even the strongest evaluated MLLM achieving less than 60\% accuracy, while \methodName{} improves MPAR accuracy by 13.38 percentage points over its base model. Our dataset is available at:
\textcolor{magenta}{https://github.com/MindIntLab-HFUT/AffToM}.
\end{abstract}

\section{Introduction}
\label{sec:intro}

Recent Multimodal Large Language Models (MLLMs) have demonstrated strong capabilities in visually grounded affective perception, including facial expression recognition, emotion description, and action understanding~\citep{lian2025affectgpt,cheng2024emotion,li2025emoverse}.
However, recognizing an apparent emotion is fundamentally different from understanding it within the person's broader cognitive and social context, as the same expression, such as a smile, may signal happiness, relief, embarrassment, concealment, or an attempt to reassure others.
Genuine emotional understanding requires reasoning beyond perceptual appearance to determine what an individual knows, expects, and believes, how they interpret the situation, and how they understand the mental states of others.

This capability is commonly referred to as \textit{affective Theory of Mind} (ToM), namely the ability to infer emotional states from the latent cognitive representations that shape them.
In realistic social interactions, emotions are not direct and universal responses to observable events but emerge from how individuals appraise those events under their beliefs, goals, expectations, and social relationships~\citep{premack1978does,lillard1993pretend,qu2015teachers}.
Consequently, the same event may evoke different emotions in different participants, while perceptually similar reactions may reflect fundamentally different mental states.
The challenge becomes greater in multi-party interactions, where an individual's emotion may depend not only on their own beliefs but also on what they believe another participant knows, intends, or feels about someone else.
For instance, a person may feel relieved because they believe a misunderstanding remains unnoticed, embarrassed because they infer that another participant has recognized their concealed intention, or guilty because they realize that one participant now understands that a third participant feels betrayed.
In fact, to evaluate machine ToM, a growing body of benchmarks has been proposed~\citep{chen-etal-2024-tombench,he2023hitombenchmarkevaluatinghigherorder,xu-etal-2024-opentom,jin2024mmtomqamultimodaltheorymind,shi2025mumatommultimodalmultiagenttheory}.

\begin{figure*}[!t]
    \centering
    \includegraphics[width=1.0\linewidth]{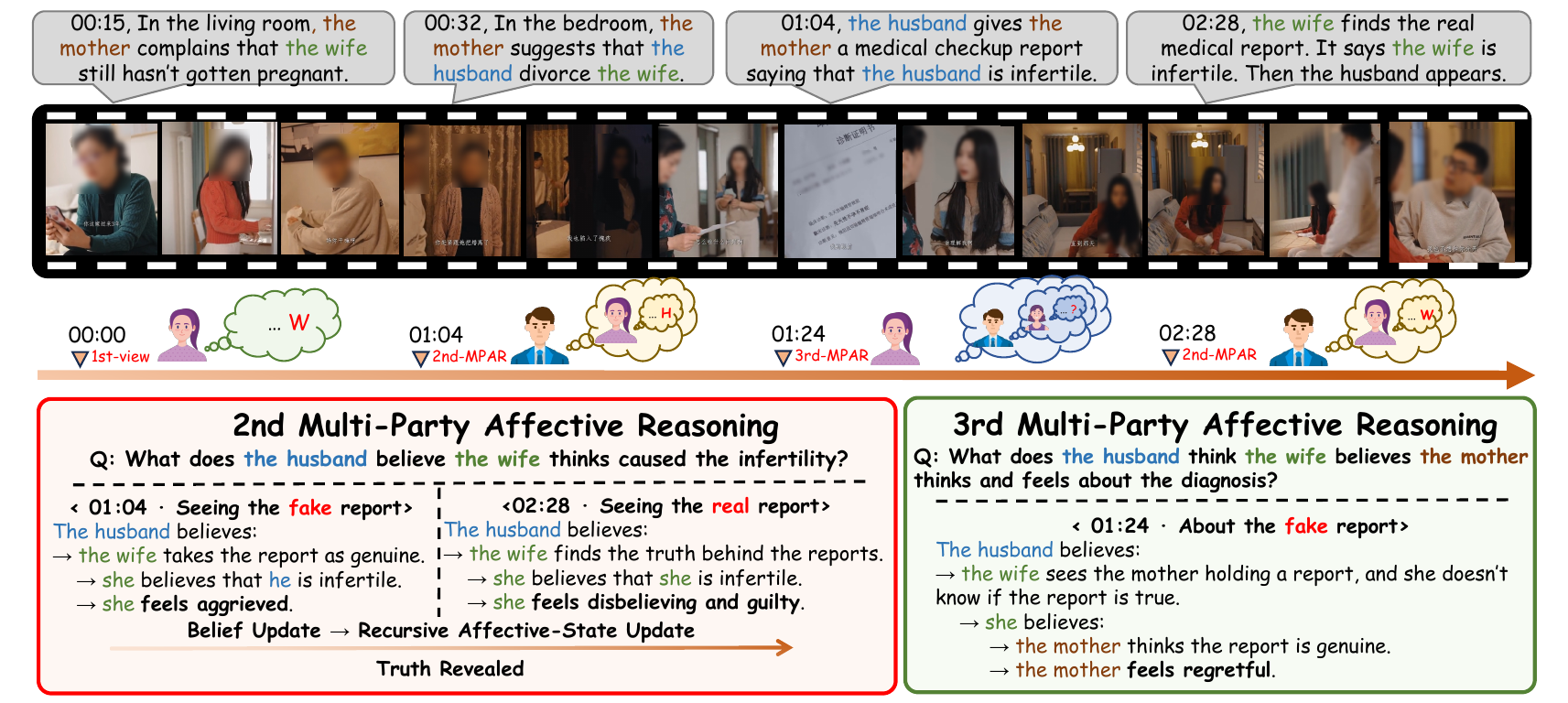}
    \caption{An {\benchmarkName} example showing multi-party affective reasoning and emotion shifts under information asymmetry.}
    \label{fig:introduction}
\end{figure*}

Despite this progress, existing benchmarks leave two closely related challenges unresolved.
First, existing higher-order ToM and affective reasoning capabilities are largely studied in isolation.
Higher-order benchmarks focus on recursively nested epistemic states, whereas affective benchmarks typically predict each participant's emotion independently, failing to capture cross-participant affective dependencies in which one person's emotion is shaped by what they believe others know, believe, or feel.
Second, current benchmarks provide only limited control over information asymmetry, recursion depth, and temporal belief updates, making it difficult to distinguish genuine higher-order affective reasoning from reliance on superficial perception.
These limitations motivate the study of \textbf{Multi-party Affective Reasoning (MPAR)}, which evaluates affective states whose correct interpretation depends on recursively nested mental representations distributed across multiple participants.
Within an ongoing social interaction, an MPAR question asks how another person believes the target character feels before the character's emotional response becomes observable.
As illustrated in Figure~\ref{fig:introduction}, the husband's affective inference changes as his information state evolves.
After the wife observes a fabricated medical report, he infers that she mistakenly believes he is infertile and interprets her affective state as feeling wronged.
When he later observes her discovering the authentic report, he updates his representation of her belief and instead infers disbelief and guilt.
Although the participants and broader situation remain largely unchanged, the correct affective inference changes because the husband's representation of what the wife knows and believes has evolved.
Therefore, we introduce {\benchmarkName}, a video-based benchmark for affective ToM that evaluates latent emotion inference from characters' evolving cognitive representations.
It contains 1,922 QA pairs from 258 videos for benchmark evaluation, complemented by 19,759 QA pairs from 2,291 videos for model training.
\begin{table*}[!htbp]
    \centering
    \scriptsize
    \caption{Comparison with representative social video and Theory-of-Mind benchmarks. 
Temporal Belief Update denotes explicit belief changes after new information, while Recursive Perspective denotes second-order or higher perspective-recursive QA.}
    \label{tab:benchmark_comparison}
    \setlength{\tabcolsep}{5pt}
    \renewcommand{\arraystretch}{1.15}
    \resizebox{0.99\linewidth}{!}{
    \begin{tabular}{l|c|cccc|cc|c}
        \hline
        \multirow{2}{*}{Benchmark}
        & \multirow{2}{*}{Real-World}
        & \multicolumn{4}{c|}{Mental State}
        & \multirow{2}{*}{\makecell{Temporal Belief Update}}
        & \multirow{2}{*}{\makecell{Recursive Perspective}}
        & \multirow{2}{*}{\makecell{Dataset Format}} \\
        \cline{3-6}
        &
        & Emotion
        & Belief
        & Intention
        & Desire
        &
        &
        & \\
        \hline
        Video-MME~\citep{fu2025videommefirstevercomprehensiveevaluation}
        & \cmark
        & \xmark & \xmark & \xmark & \xmark
        & \xmark & \xmark
        & MCQ \\

        Hi-ToM~\citep{he2023hitombenchmarkevaluatinghigherorder}
        & \xmark
        & \xmark & \cmark & \xmark & \xmark
        & \xmark & \cmark
        & MCQ \\

        Social-IQ~\citep{zadeh2019socialiq}
        & \cmark
        & \cmark & \cmark & \cmark & \cmark
        & \xmark & \xmark
        & MCQ \\

        MMToM-QA~\citep{jin2024mmtomqamultimodaltheorymind}
        & \xmark
        & \xmark & \cmark & \cmark & \xmark
        & \xmark & \xmark
        & MCQ \\
        
        MOMENTS~\citep{villa-cueva-etal-2025-moments}
        & \cmark
        & \cmark & \cmark & \cmark & \cmark
        & \xmark & \xmark
        & MCQ \\

        R$^3$-Bench~\citep{niu2026read}
        & \cmark
        & \cmark & \cmark & \cmark & \cmark
        & \xmark & \xmark
        & MCQ \\
        \hline
        \textbf{{\benchmarkName}}
        & \cmark
        & \cmark & \cmark & \cmark & \cmark
        & \cmark & \cmark
        & MCQ + Open-ended \\
        \hline
    \end{tabular}
    }
    \vspace{-1.5em}
\end{table*}
{\benchmarkName} evaluates two complementary capabilities: \textbf{Individual Affective Reasoning (IAR)}, which requires models to infer an individual's affective state by integrating observable events with situational context, and \textbf{MPAR}, which further requires models to recursively track how participants represent one another's beliefs and emotions and use these nested representations to infer affective states.
We develop an event-driven multi-agent question-generation pipeline grounded in dynamic character interaction graphs that explicitly represent event chronology, interpersonal relations, and mental-state transitions.
These structured graphs enable controlled generation of questions with different participant configurations, information asymmetries, temporal update depths, and levels of mental-state recursion.
Role-specialized agents generate IAR and MPAR questions from the graphs, after which the resulting samples are refined through an iterative dual-criteria evolution process that jointly improves logical validity and distractor discriminability, followed by human validation.
These findings motivate us to examine whether targeted training can improve belief-conditioned emotion inference from others' perspectives.
We introduce \methodName{}, a two-stage post-training framework that applies supervised fine-tuning (SFT) followed by perspective-aware rubric-guided reinforcement learning.
Comprehensive evaluations of state-of-the-art MLLMs under both multiple-choice and open-ended settings show that MPAR remains particularly challenging, with even the strongest evaluated model achieving less than 60\% accuracy.
Meanwhile, \methodName{} improves MPAR accuracy by 13 percentage points over its base model.

\begin{figure*}[!t]
    \centering
    \includegraphics[width=1.0\linewidth
    ]{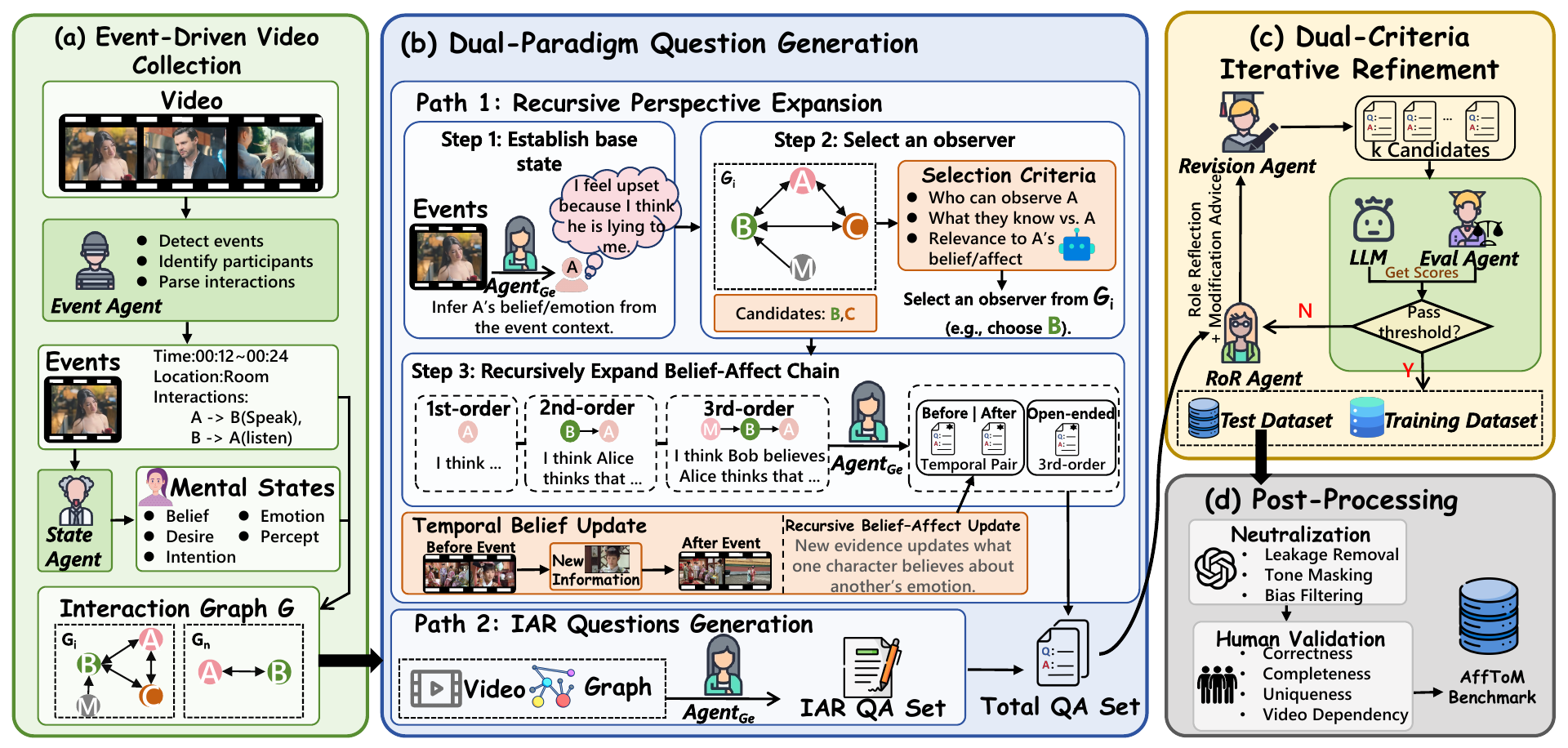}
    \caption{
        Overview of the dataset construction pipeline, including event-driven video collection, question generation, data evolution, and manual evaluation.
    }
    \label{fig:frameWork}
\end{figure*}

\section{Related Work}

\noindent \textbf{Theory of Mind.} Theory of Mind (ToM) refers to the capacity to infer others' mental states, including beliefs, intentions, and emotions~\citep{premack1978does,decety2004functional}.
Although ToM is frequently operationalized through belief and intention reasoning, psychological theories increasingly recognize emotion as an important component of mental-state representation. Since attributing emotion depends critically on tracking others’ mental states~\cite{qu2015teachers}, emotion understanding should be regarded as a fundamental aspect of ToM rather than a peripheral add-on.
Early research on ToM has primarily been conducted in conversational or game-based settings~\cite{liu2024interintentinvestigatingsocialintelligence}.
Text-based benchmarks evaluate belief tracking, false-belief reasoning, and recursive perspective-taking~\citep{le2019revisiting,kim2023fantombenchmarkstresstestingmachine,he2023hitombenchmarkevaluatinghigherorder,xu-etal-2024-opentom,chen-etal-2024-tombench}, while recent multimodal benchmarks extend ToM to visually grounded and interactive settings.
MMToM-QA and MuMA-ToM study beliefs and goals in multi-agent environments~\citep{jin2024mmtomqamultimodaltheorymind,shi2025mumatommultimodalmultiagenttheory}, whereas MOMENTS, Social-IQ 2.0, and IntentQA evaluate broader social reasoning abilities from real-world videos~\citep{villa-cueva-etal-2025-moments,siq2,li2023intentqa}.
More recent work further explores mental-physical causal reasoning and multimodal emotion reasoning~\citep{niu2026read,luo2026unveiling}.
Despite this progress, emotion inference through nested mental states and its evolution remains underexplored.

\noindent \textbf{Multimodal Question Answering.} 
LLMs have achieved remarkable progress in question answering across a wide range of domains~\cite{yang2025qwen3,team2026kimi,hu2024psycollm,dai2026psyche,xu2025multiagentesc,dou2026baichuan}.
With the integration of visual inputs, multimodal QA has further extended language-based reasoning to visually grounded scenarios.
Early works in multimodal QA focus on image question answering.
Subsequently, the field moves from recognition toward deeper understanding and commonsense modeling. 
Benchmarks such as PerceptionTest~\cite{pătrăucean2023perceptiontestdiagnosticbenchmark} and HourVideo~\cite{chandrasegaran2024hourvideo1hourvideolanguageunderstanding} begin to emphasize temporal integration, long-video context, and limited forms of causal understanding, requiring models to go beyond perception and answer “what happened?” and “why did it happen?” 
On this basis, multimodal QA has increasingly expanded into the affective domain, where models must jointly interpret facial expressions, prosody, language, and broader social context. 
Early benchmarks mainly focus on sentiment or emotion recognition, emphasizing polarity, intensity, or categorical emotion prediction, as exemplified by MOSI~\cite{zadeh2016mosimultimodalcorpussentiment},  CMU-MOSEI~\cite{bagher-zadeh-etal-2018-multimodal}, MTMEUR~\cite{hu2025beyond}, CEDAR~\cite{dai2026tears}, MELD~\cite{poria-etal-2019-meld}, and MMHBench~\cite{hu2026mmhbench}.
More recent datasets have moved beyond closed-set labeling toward more explainable and reasoning-oriented affect understanding. 
EMER~\cite{lian2025affectgpt} and DEEMO~\cite{li2025deemo} introduce explanatory rationales, OV-MER~\cite{lian2024ov} relaxes fixed label spaces through open-vocabulary emotion prediction, and MME-Emotion~\cite{zhang2025mme} further evaluates both emotional understanding and causal reasoning in a unified and interpretable QA-style setting.
Table~\ref{tab:benchmark_comparison} compares \benchmarkName{} with representative social video and ToM benchmarks.
\benchmarkName{} brings together real-world video grounding, temporal belief updates, and recursive perspective-taking to evaluate how emotions depend on participants' evolving beliefs about one another.

\section{{\benchmarkName}}
\label{sec:afftom}
In this section, we detail the systematic construction of the {\benchmarkName} benchmark.
We introduce a multi-stage data generation pipeline driven by a collaborative multi-agent framework, with the overall structure shown in Figure \ref{fig:frameWork}.
This systematic approach sequentially consists of event-driven video curation, dual-paradigm question generation enhanced by a social interaction graph, dual-criteria-guided data evolution, and comprehensive post-processing alongside expert validation.

\subsection{Event-Driven Video Collection}
We collect 1,504 raw video candidates from public video-sharing platforms, including Douyin, TikTok, and YouTube.
Collection details are provided in Appendix \ref{app:dataset_statistics}.
These videos span diverse social scenarios, thereby providing rich multimodal signals for modeling complex human behaviors.
To systematically parse the semantic content of these candidate videos, we employ Seed-1.8~\citep{seed2026seed18modelcardgeneralized} to extract fine-grained event metadata.
For video $v$, the event extraction is defined as:
\begin{equation}
    \mathcal{E} = \text{Agent}_{\text{eve}}(v),
\end{equation}
where $\mathcal{E} = \{e_{1}, e_{2}, \dots, e_{N}\}$ denotes the set of events extracted from the video.
Each event $e_{i}$ consists of its spatiotemporal grounding—comprising the timestamp, location, and involved individuals.
To capture multi-party social dynamics for affective ToM reasoning, we retain videos with multiple characters connected by explicit directed interaction pairs.
This filtering yields 308 videos, for which we extract character-level cognitive and affective attributes:
\begin{equation}
    \mathcal{S} = \text{Agent}_{\text{state}}(v, \mathcal{E}).
\end{equation}
Here, $\mathcal{S} = \{s_{1}, s_{2}, \dots, s_{M}\}$ denotes the set of individual-level socio-cognitive states, encompassing identity traits, interpersonal relationships, and dynamically evolving mental states of the characters throughout the video, where $s_i$ corresponds to the $i$-th character $c_i$.
Using events $\mathcal{E}$ and individual-level information $\mathcal{S}$, we construct a dynamic character interaction graph $G = \{G_{1}, G_{2}, \dots, G_{N}\}$, where nodes represent the characters and edges encode their evolving behavioral relationships.

\subsection{Dual-Paradigm Question Generation}
After constructing the dynamic character interaction graph $G$, we design two complementary question generation paths to construct affective ToM reasoning tasks.

\noindent \textbf{Recursive Perspective Expansion.}
The first paradigm generates Multi-party Affective Reasoning (MPAR) questions through recursive perspective expansion.
For each extracted event, we establish a base-level affective state or belief for a target individual, grounded in the event context:
\begin{equation}
    \mathcal{Q}^{(1)} = \Phi_{\text{init}}(\mathcal{E}, v),
\end{equation}
$\mathcal{Q}^{(1)} = \{q_{1}^{1},q_{2}^{1},\dots, q_{N}^{1}\}$ serves as the foundational belief questions from which more complex reasoning structures are derived.
To construct MPAR questions, we progressively introduce additional individuals from the interaction graph $G$ and recursively integrate their respective perspectives into the reasoning process.
Second-order questions introduce a second individual and require reasoning about one person's belief concerning another's mental or emotional state.
Third and higher orders add further parties and their perspectives.
Rather than merely nesting beliefs, the questions capture how attitudes, expectations, and misunderstandings shape emotions within the shared interaction, evaluating dynamic, interaction-driven affective reasoning beyond isolated emotion recognition.
We formally define this recursive generation of higher-order cognitive states as:
\begin{equation}
    (\mathcal{Q}^{(i+1)},\mathcal{A}^{(i+1)}) = \Phi_{\text{recur}}(\mathcal{Q}^{(i)}, G_{i}, v),
\end{equation} 
where $\mathcal{Q}^{(i)}$ denotes the $i$-th order question and $\mathcal{A}^{(i+1)}$ denotes the corresponding answer.
$\Phi_{\text{recur}}$ is the recursive integration function that expands the current reasoning structure by incorporating the perspective of a newly introduced character $c_k$, using the corresponding event and character information encoded in $G_{i}$, together with the video $v$.

To quantify the behavioral effect of the interaction, we formulate some higher-order queries as temporal pairs spanning across the event timeline.
Specifically, we generate coupled questions querying the nested affective states at both a pre-event timestamp $t_{\text{pre}}$ and a post-event timestamp $t_{\text{post}}$.
This contrasting design forces models to explicitly trace the temporal evolution of emotional states and measure the causal impact of the specific event on resolving or exacerbating interpersonal misunderstandings.
Upon deriving the definitive question-answer pairs $(\mathcal{Q}, \mathcal{A})$, we synthesize a set of semantically plausible distractors to construct the final multiple-choice questions (MCQs).
Furthermore, to complement this standard discriminative evaluation, we introduce open-ended MPAR questions specifically designed to assess deeply nested higher-order affective reasoning.
An event-driven MPAR generation example is provided in Appendix~\ref{app:mpar_generation_example}.

\noindent \textbf{Individual Affective Reasoning Generation.}
Complementing multi-party relational reasoning, the second paradigm is designed to formulate individual affective reasoning (IAR) questions that probe the internal states and latent characteristics of individual characters within emotionally rich social contexts.
Specifically, this track covers six dimensions of affective and behavioral understanding, including Preference Prediction, Integrated Mental-State Reasoning, Strategic Perspective-Taking, Hidden Intention, Personality Reasoning, and Counterfactual Reasoning.
For this type, we generate the question through a question generation agent with video $v$ and the graph $G$ as inputs, guided by a curated set of few-shot exemplars derived from human-validated seed questions.

\subsection{Dual-Criteria Iterative Refinement}

To improve cognitive depth and linguistic quality, our iterative pipeline combines role-guided reflection with dual-criteria evaluation.
The Role Reflection Agent ($\text{Agent}_{\text{RoR}}$) examines each generated question from the event's different roles, identifying improvements, ambiguities, and alternative viewpoints grounded in the interaction structure.
Using this feedback, the Revision Agent ($\text{Agent}_{\text{rev}}$) refines the initial drafts into a diverse candidate pool $\mathcal{Q}_{\text{cand}}$.
We then select high-quality questions from $\mathcal{Q}_{\text{cand}}$ using two criteria: \textit{Comparative Rationality} and \textit{Distractor Unbiasedness}.

\noindent \textbf{Criterion 1: Comparative Rationality ($S_{\text{rank}}$).}
Most existing data generation pipelines adopt scoring-based selection strategies to identify high-quality candidates from a pool of generated samples{~\citep{li2025add, agarwal-etal-2025-promptwizard}.}
However, relying on standard absolute scoring or uncalibrated list-wise ranking often introduces severe vulnerabilities, such as verbosity, self-enhancement, and position biases~\citep{zheng2023judgingllmasajudgemtbenchchatbot}.
To rigorously isolate the highest-quality questions and circumvent these biases, we employ a robust pairwise comparison mechanism benchmarked against human expertise.
Specifically, we prompt an evaluation agent (i.e., $\text{Agent}_{\text{eval}}$) to conduct blind pairwise evaluations between a generated candidate $q_c \in \mathcal{Q}_{\text{cand}}$ and a sampled human-authored reference question $q_h \in \mathcal{Q}_{\text{human}}$ from the seed pool.
$\text{Agent}_{\text{eval}}$ judges the candidate superior, equivalent, or inferior to the human reference based on logical coherence, contextual alignment, and affective depth.
The comparative rationality score is the expected win rate over the human references:
\begin{multline}
S_{\text{rank}}(q_c)
= \frac{1}{|\mathcal{Q}_{\text{human}}|}
  \sum_{q_h \in \mathcal{Q}_{\text{human}}}
\\
\Bigl(
  \mathbb{I}(q_c \succ q_h)
  + \tfrac{1}{2}\mathbb{I}(q_c \sim q_h)
\Bigr).
\end{multline}
where $\mathbb{I}(\cdot)$ is the indicator function, $q_c \succ q_h$ denotes that the candidate outperforms the human reference, and $q_c \sim q_h$ denotes a tie in quality.
This pairwise formulation, bounded between 0 and 1, mitigates list-wise position bias and ensures that a high score reflects a quality level authentically comparable to, or explicitly exceeding, expert human annotations.

\noindent \textbf{Criterion 2: Distractor Unbiasedness ($S_{\text{bias}}$).}
In multimodal benchmarks, models often exploit linguistic priors to guess the correct answer without processing the visual input.
To ensure our distractors (i.e., incorrect options) are plausible and effectively penalize blind guessing, we evaluate the candidate options $\mathcal{O} = \{A, B, C, D\}$ under a video-blind setting.
Given only the textual context, we estimate the prior probability distribution over the options using an LLM.
The LLM produces output logits for each candidate option, which are converted into a probability distribution $P(o \mid \text{text-only})$ via a softmax function.
An unbiased question should compel the text-only model into a state of maximum uncertainty, yielding a uniform distribution across all options.
We formally quantify this uncertainty using Normalized Shannon Entropy:
\begin{equation}
S_{\text{bias}}(q_c) = -\frac{1}{\log |\mathcal{O}|} \sum_{o \in \mathcal{O}} p_o \log p_o,
\end{equation}
where $p_o = P(o \mid \text{text-only})$ denotes the probability assigned to option $o$ by the LLM given only the textual context, without access to the video.
If the model is highly confident in a single option without seeing the video, $S_{\text{bias}}$ yields a low score, indicating that the distractors are ineffective or the question is textually biased.

\noindent \textbf{Iterative Evolution and Termination.}
We combine the two criteria into a unified score for iterative candidate refinement:
\begin{equation}
    S_{\text{total}}(q_c) = \lambda S_{\text{rank}}(q_c) + (1 - \lambda) S_{\text{bias}}(q_c),
\end{equation}
where $\lambda$ controls the trade-off between question quality and distractor robustness.
If a generated candidate meets a predefined acceptance threshold $\tau$, it is accepted as a high-quality question.
Otherwise, the generation process enters the next evolution cycle.
This iterative reflection–revision–ranking process continues until a candidate question satisfies the criterion.
To prevent unbounded iterations, we impose a predefined maximum number of refinement rounds $T_{\text{max}}$.
Let $s_t = S_{\text{total}}(q_c^{(t)})$ denote the candidate's score at iteration $t$.
The selection and routing process is defined as:
\begin{equation}
q^{*} = \begin{cases}
q_c^{(t)} & s_t \ge \tau, \\
\Phi_{\text{evolve}}\bigl(q_c^{(t)}\bigr) & s_t < \tau,\ t < T_{\text{max}}, \\
\emptyset & t \ge T_{\text{max}}.
\end{cases}
\end{equation}
Here, $\Phi_{\text{evolve}}$ performs role-guided reflection and revision.
The reflection agent $\text{Agent}_{\text{RoR}}$ uses the interaction graph $G_i$ to assess the candidate $q_c^{(t)}$.
The revision agent $\text{Agent}_{\text{rev}}$ then revises the candidate using this feedback.
The threshold $\tau$ controls acceptance, while $T_{\text{max}}$ limits the number of refinement cycles.

\subsection{Post-Processing and Human Validation}
We additionally apply the same generation pipeline to a separate pool of 2,291 training videos, yielding 19,759 QA samples.
The test set undergoes two additional steps: linguistic post-processing and human validation.
First, to reduce linguistic shortcuts, we apply linguistic neutralization to the questions and options retained after iterative refinement.
We remove explicit emotional polarity and soften overly deterministic phrasing, for example by rewriting ``react negatively'' as ``show a particular emotional reaction''.
This encourages models to reason about multimodal context rather than rely on textual cues.
We manually validate the questions against the source videos (see validation details in Appendix~\ref{app:human_validation}).
Through this multi-stage filtering and validation process, we obtain a benchmark containing {\questionNum} high-quality questions grounded in {\videoNum} videos.
Detailed statistics are provided in Appendix~\ref{app:dataset_statistics}.

\begin{figure}[!t]
    \centering
    \includegraphics[width=\linewidth]{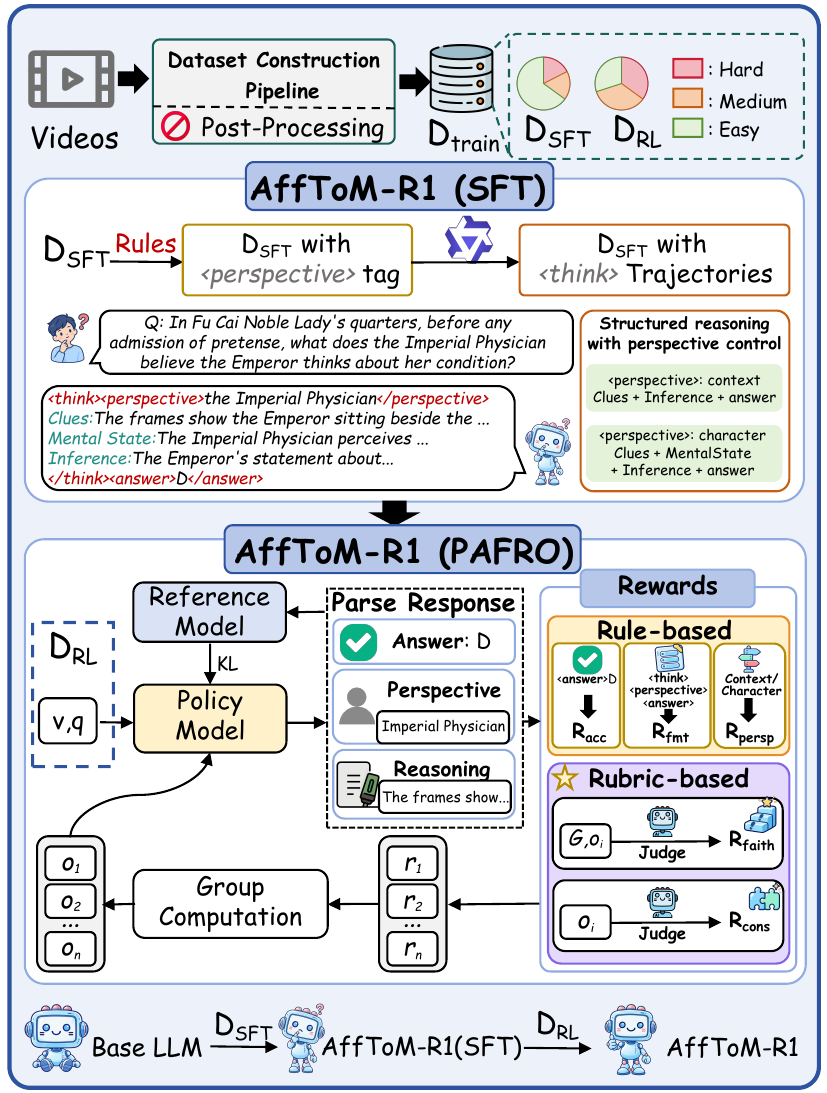}
    \caption{Overview of \methodName{}.}
    \label{fig:grpo_framework}
\end{figure}

\section{\methodName{}}
\label{sec:method}

In this section, we present \methodName{}.
As illustrated in Figure~\ref{fig:grpo_framework}, SFT trains the model on structured reasoning trajectories, while Perspective-Aware Faithfulness Reasoning Optimization (\grpoName{}) further optimizes it with perspective-aware rubric-based rewards.
We construct a 20k-sample training pool and split it into $\mathcal{D}_{\mathrm{SFT}}$ for SFT and $\mathcal{D}_{\mathrm{RL}}$ for \grpoName{} based on difficulty estimated from the pretrained backbone.
Further training details are provided in Appendix~\ref{app:training_settings}.

\subsection{Supervised Fine-Tuning}
\label{sec:routed_sft}

We first perform supervised fine-tuning (SFT) on perspective-structured reasoning trajectories that connect video evidence to the target answer.
Each target response consists of a reasoning block enclosed by \texttt{<think>}\allowbreak\texttt{</think>} and a final answer enclosed by \texttt{<answer>}\allowbreak\texttt{</answer>} (see Appendix \ref{app:sft_construction} for more details).
The reasoning block begins with a \texttt{<perspective>}\allowbreak\texttt{</perspective>} field that explicitly identifies the viewpoint from which the question is resolved to guide later reasoning.

\noindent\textbf{Perspective-structured reasoning.}
We adopt two reasoning structures depending on whether the question requires character-specific mental-state inference.
For the \emph{character-centered} structure, the trajectory follows \emph{Clues} $\rightarrow$ \emph{Mental State} $\rightarrow$ \emph{Inference}, explicitly interpreting the observed evidence through the target character's information and mental state before deriving the answer.
For the \emph{context-centered} structure, the trajectory follows \emph{Clues} $\rightarrow$ \emph{Inference} when explicit character-level mental-state modeling is unnecessary.
All MPAR questions adopt the \emph{character-centered} structure, while the remaining questions are assigned an appropriate structure with LLM assistance.
We generate candidate trajectories accordingly and retain only those whose final answers match the reference answers, yielding $\mathcal{D}_{\mathrm{SFT}}$ for SFT.

\subsection{Perspective-Aware Faithfulness Reasoning Optimization}
\label{sec:reasoning_grpo}

\grpoName{} further optimizes the SFT-trained model using rubric-based GRPO, as illustrated in Figure~\ref{fig:grpo_framework}.
Details of GRPO are provided in Appendix~\ref{app:grpo_optimization}.
\begin{table*}[!t]
\centering
\small
\setlength{\tabcolsep}{4pt}
\renewcommand{\arraystretch}{1.12}
\caption{
Performance comparison of different MLLMs. 
Human performance is evaluated on 250 questions randomly sampled from {\benchmarkName}  (see Appendix \ref{app:human_baseline} for more details).
}
\label{tab:main_results}
\resizebox{0.99\linewidth}{!}{
\begin{tabular}{
l
!{\vrule width 0.5pt}
c
!{\vrule width 0.5pt}
cccccccc
!{\vrule width 0.5pt}
ccc
}

\toprule

\multirow{2}{*}{\textbf{Models}}
& \multirow{2}{*}{\textbf{Params}}
& \multicolumn{8}{c!{\vrule width 0.5pt}}{\textbf{Multiple-Choice Question}}
& \multicolumn{3}{c}{\textbf{Open-ended}} \\

\cmidrule(lr){3-10}
\cmidrule(lr){11-13}

&
& \textbf{PP}
& \textbf{IMSR}
& \textbf{SPT}
& \textbf{HI}
& \textbf{PR}
& \textbf{CR}
& \textbf{MPAR}
& \textbf{Avg.}
& \textbf{EC}
& \textbf{FC}
& \textbf{Avg.}
\\

\midrule

\rowcolor{gray!20}
\multicolumn{13}{c}{\textit{Open-Source MLLMs}} \\

\midrule

\rowcolor{blue!5}
Ministral-3-3B
& 3B
& 31.02 & 39.81 & 43.15 & 27.49 & 38.89 & 33.33 & 30.07 & 33.26
& 1.64 & 1.48 & 1.56 \\

\rowcolor{blue!5}
Gemma-3-4B
& 4B
& 40.11 & 52.43 & 43.15 & 39.77 & 42.78 & 37.85 & 29.67 & 37.43
& 1.62 & 1.26 & 1.44 \\

\rowcolor{blue!5}
LLaVA-OneVision-1.5
& 4B
& 47.59 & 63.11 & 60.96 & 47.95 & 45.00 & 48.59 & \textbf{39.34} & 46.87
& 1.82 & 1.51 & 1.67 \\

\rowcolor{blue!5}
Qwen3-VL-4B-Instruct
& 4B
& 50.00 & \textbf{63.41} & 51.37 & 53.80 & 44.44 & 46.89 & 36.42 & 45.50
& \textbf{1.88} & \textbf{1.57} & \textbf{1.73} \\

\rowcolor{blue!5}
Qwen3.5-4B
& 4B
& \textbf{52.13} & 61.46 & \textbf{66.44} & \textbf{55.56} & \textbf{54.44} & \textbf{58.19} & 37.22 & \textbf{49.29}
& 1.77 & 1.50 & 1.64 \\

\midrule

\rowcolor{blue!5}
Ministral-3-8B
& 8B
& 44.92 & 50.00 & 48.63 & 45.03 & 48.89 & 40.68 & 32.05 & 40.45
& 1.79 & 1.60 & 1.70 \\

\rowcolor{blue!5}
LLaVA-OneVision-Qwen2
& 7B
& 40.64 & 50.00 & 52.74 & 47.95 & 41.11 & 44.63 & 33.91 & 41.00
& 1.85 & 1.48 & 1.67 \\

\rowcolor{blue!5}
InternVL3.5-8B
& 8B
& 50.27 & 52.91 & 58.22 & 50.29 & 53.33 & 54.24 & 32.32 & 44.46
& 1.88 & 1.60 & 1.74 \\

\rowcolor{blue!5}
MiniCPM-V-2.6
& 7B
& 44.39 & 61.65 & 61.64 & 49.71 & 47.22 & 53.67 & 36.03 & 45.94
& 1.76 & 1.58 & 1.67 \\

\rowcolor{blue!5}
Qwen2.5-VL-7B-Instruct
& 7B
& 52.41 & 58.74 & 58.22 & 50.29 & 52.22 & 53.67 & 36.03 & 46.71
& 1.85 & 1.58 & 1.72 \\

\rowcolor{blue!5}
LLaVA-OneVision-1.5
& 8B
& 44.39 & \textbf{67.48} & 63.01 & 52.63 & 56.67 & 57.63 & 33.77 & 47.37
& \textbf{1.96} & 1.61 & 1.79 \\

\rowcolor{blue!5}
MiniCPM-V-4.5
& 8B
& 47.59 & 62.62 & 67.12 & 60.23 & 56.11 & 55.93 & 39.60 & 50.38
& 1.49 & 1.31 & 1.40 \\

\rowcolor{blue!5}
Qwen3-VL-8B-Instruct
& 8B
& \textbf{59.36} & 66.99 & 58.90 & \textbf{61.99} & \textbf{59.44} & 54.80 & 39.87 & 51.92
& 1.92 & \textbf{1.70} & \textbf{1.81} \\

\rowcolor{blue!5}
Qwen3.5-9B
& 9B
& 57.22 & \textbf{67.48} & \textbf{68.49} & \textbf{61.99} & 57.22 & \textbf{64.97} & \textbf{40.40} & \textbf{53.51}
& 1.79 & 1.52 & 1.66 \\

\midrule

\rowcolor{blue!5}
Gemma-3-27B
& 27B
& 50.27 & 58.74 & 45.21 & 55.56 & 53.89 & 45.20 & 30.07 & 42.81
& 1.87 & 1.45 & 1.66 \\

\rowcolor{blue!5}
InternVL3.5-14B
& 14B
& 49.20 & 54.37 & 63.70 & 49.71 & 48.89 & 51.41 & 34.04 & 44.90
& 1.82 & 1.62 & 1.72 \\

\rowcolor{blue!5}
InternVL3.5-38B
& 38B
& 52.94 & 58.25 & 64.38 & 46.78 & 53.33 & 59.32 & 35.36 & 47.26
& 1.99 & 1.76 & 1.88 \\

\rowcolor{blue!5}
Qwen2.5-VL-72B-Instruct
& 72B
& 54.55 & 72.33 & 65.75 & 61.40 & 61.67 & \textbf{66.10} & 44.24 & 55.65
& 2.22 & 1.95 & 2.09 \\

\rowcolor{blue!5}
Qwen3-VL-32B-Instruct
& 32B
& 61.50 & \textbf{77.67} & 66.44 & 63.16 & \textbf{64.44} & 64.41 & 40.79 & 55.87
& 2.39 & 1.96 & 2.18 \\

\rowcolor{blue!5}
Qwen3.5-27B
& 27B
& \textbf{67.91} & 75.73 & \textbf{72.60} & \textbf{69.01} & 63.89 & \textbf{66.10} & \textbf{45.03} & \textbf{59.22}
& \textbf{2.45} & \textbf{2.23} & \textbf{2.34} \\

\midrule

\rowcolor{gray!20}
\multicolumn{13}{c}{\textit{Closed-Source MLLMs}} \\

\midrule

\rowcolor{blue!5}
Seed-2.0-lite
& --
& 65.78 & 72.82 & \textbf{71.23} & 70.76 & 62.78 & 66.10 & 44.24 & 58.29
& 2.02 & 1.88 & 1.95 \\

\rowcolor{blue!5}
GPT-5.5
& --
& \textbf{71.28} & \textbf{76.59} & 68.49 & \textbf{71.93} & \textbf{66.67} & \textbf{74.01} & \textbf{52.05} & \textbf{63.56}
& \textbf{3.06} & \textbf{2.99} & \textbf{3.03} \\

\midrule

\rowcolor{gray!20}
\multicolumn{13}{c}{\textit{Ours}} \\

\midrule

\rowcolor{blue!5}
\methodName{} (Qwen3.5-4B)
& 4B
& 54.79 & 73.17 & 72.60 & 63.16 & 63.33 & 66.67 & 50.60 & 59.33
& 2.17 & 1.89 & 2.03 \\

\rowcolor{blue!5}
Human Baseline
& --
& 91.30 & 92.59 & 90.48 & 95.00 & 92.00 & 91.67 & 84.55 & 88.80
& -- & -- & -- \\

\bottomrule

\end{tabular}}
\end{table*}

\noindent\textbf{Reward design.}
Answer correctness alone does not reveal whether the model adopts the appropriate viewpoint or supports its conclusion with video evidence.
We therefore combine rule-based checks of answer correctness ($R_{\mathrm{acc}}$), response format ($R_{\mathrm{fmt}}$), and perspective alignment ($R_{\mathrm{persp}}$) with two reasoning rubrics.
Mental-state faithfulness ($R_{\mathrm{faith}}$) checks the response's claims about beliefs, intentions, and emotions by comparing them with the dynamic character interaction graph $G$ provided as reference, using the relevant event or whole-video scope.
For \emph{character-centered} responses, this rubric assesses claims from the declared character's viewpoint.
\emph{Context-centered} responses are assessed without a single-character viewpoint.
Evidence-answer consistency ($R_{\mathrm{cons}}$) checks whether sampled frames and subtitles support the reasoning and whether that reasoning supports the selected answer.
It does not use the reference answer, allowing the reward to distinguish evidential support from answer correctness.
The response-level reward is:
\begin{equation}
    \begin{aligned}
        r_i &= \lambda_{\mathrm{acc}}R_{\mathrm{acc}}
            + \lambda_{\mathrm{fmt}}R_{\mathrm{fmt}}
            + \lambda_{\mathrm{persp}}R_{\mathrm{persp}} \\
        &\quad + \lambda_{\mathrm{faith}}R_{\mathrm{faith}}
            + \lambda_{\mathrm{cons}}R_{\mathrm{cons}},
    \end{aligned}
    \label{eq:combined_reward}
\end{equation}
where each component is evaluated for response $y_i$ and lies in $[0,1]$.
The $\lambda$ coefficients set their relative weights.
Appendix~\ref{app:reward_definitions} provides the scoring rules for all five components.

\section{Experiments}

\subsection{Experimental Settings}

\noindent\textbf{Evaluated Models.}
We evaluate open-source MLLMs across model families and scales, including the Qwen2.5-VL series~\citep{bai2025qwen25vltechnicalreport}, the Qwen3-VL series~\citep{bai2025qwen3vltechnicalreport}, the Qwen3.5 series~\citep{qwen3.5}, the Gemma-3 series, the InternVL3.5 series~\citep{wang2025internvl35advancingopensourcemultimodal}, MiniCPM-V-2.6~\citep{yao2024minicpm}, MiniCPM-V-4.5~\citep{yu2025minicpmv45cookingefficient}, models from the Ministral-3 series, LLaVA-OneVision~\citep{li2024llavaonevisioneasyvisualtask}, and LLaVA-OneVision-1.5~\citep{an2025llavaonevision15fullyopenframework}.
We also evaluate closed-source models, including GPT-5.5 and Seed-2.0-lite~\citep{seed2026seed20modelcardintelligence}.
The evaluated configurations are listed in Table~\ref{tab:main_results}.
For the training study, we compare Qwen3.5-4B, its SFT variant, and \methodName{} on \benchmarkName{} and external benchmarks covering social reasoning and higher-order belief reasoning.

\noindent  \textbf{Evaluation Metrics.}
For MCQs, we utilize standard accuracy as the primary quantitative metric.
For open-ended queries, we further employ GPT-5-mini to score the generated responses on a standard 1-to-5 scale across Emotion Correctness (EC) and Factual Consistency (FC), where 1 indicates poor quality and 5 indicates high quality.
\textbf{EC} evaluates whether the predicted affective state accurately matches the ground-truth emotion experienced by the target character.
\textbf{FC} measures the extent to which the generated response remains faithful to the ground-truth context.
Agreement between GPT-5-mini and human scores is assessed in Appendix~\ref{app:judge_validation}.

\noindent\textbf{Inference Settings.}
We set the decoding temperature to $0$ during evaluation and use 64 uniformly sampled video frames per question unless otherwise specified.
Open-source MLLMs are deployed and evaluated on NVIDIA RTX 5090 GPUs, while closed-source models are accessed through their vendors' commercial APIs.

\subsection{Main Results}
We evaluate a diverse and representative set of MLLMs on the proposed benchmark.
The overall performance metrics are summarized in Table \ref{tab:main_results}, from which we derive several key and informative insights into current model capabilities.
First, we can observe that MPAR is consistently the most challenging MCQ category.
Even GPT-5.5, the strongest evaluated model, achieves only 52.05\% accuracy on MPAR, substantially lower than its performance on the other question types.
This gap highlights the difficulty of recursively tracking and integrating the affective states of multiple participants, exposing the limitations of current MLLMs in higher-order affective ToM reasoning.
Second, larger models generally achieve stronger performance within the same family.
This trend is observed in both MCQ and open-ended evaluations, suggesting that increased capacity is associated with better multimodal and socio-cognitive reasoning.
The Qwen3.5 series is particularly competitive, with Qwen3.5-27B achieving the highest average scores among open-source MLLMs.
Third, large-scale MLLMs yield further improvements across both evaluation formats.
GPT-5.5 achieves the best overall performance on both MCQs and open-ended questions, demonstrating stronger capabilities in answer selection and free-form affective reasoning.
\begin{figure}[t]
    \centering
    \includegraphics[width=\linewidth]{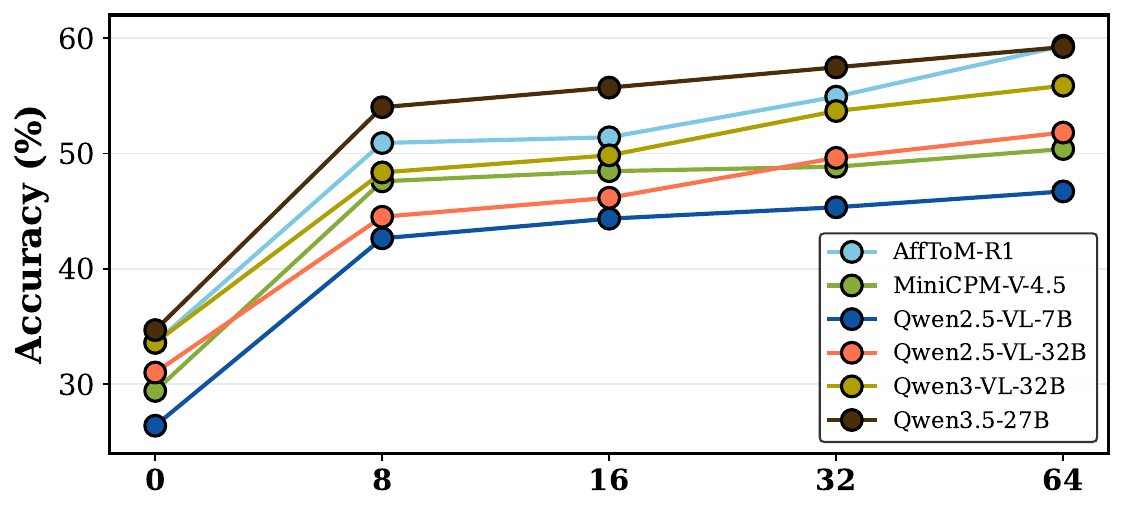}
    \caption{Accuracy across frame counts.}
    \label{fig:frame_count_accuracy}
\end{figure}
\begin{figure}[t]
    \centering
    \includegraphics[width=\linewidth]{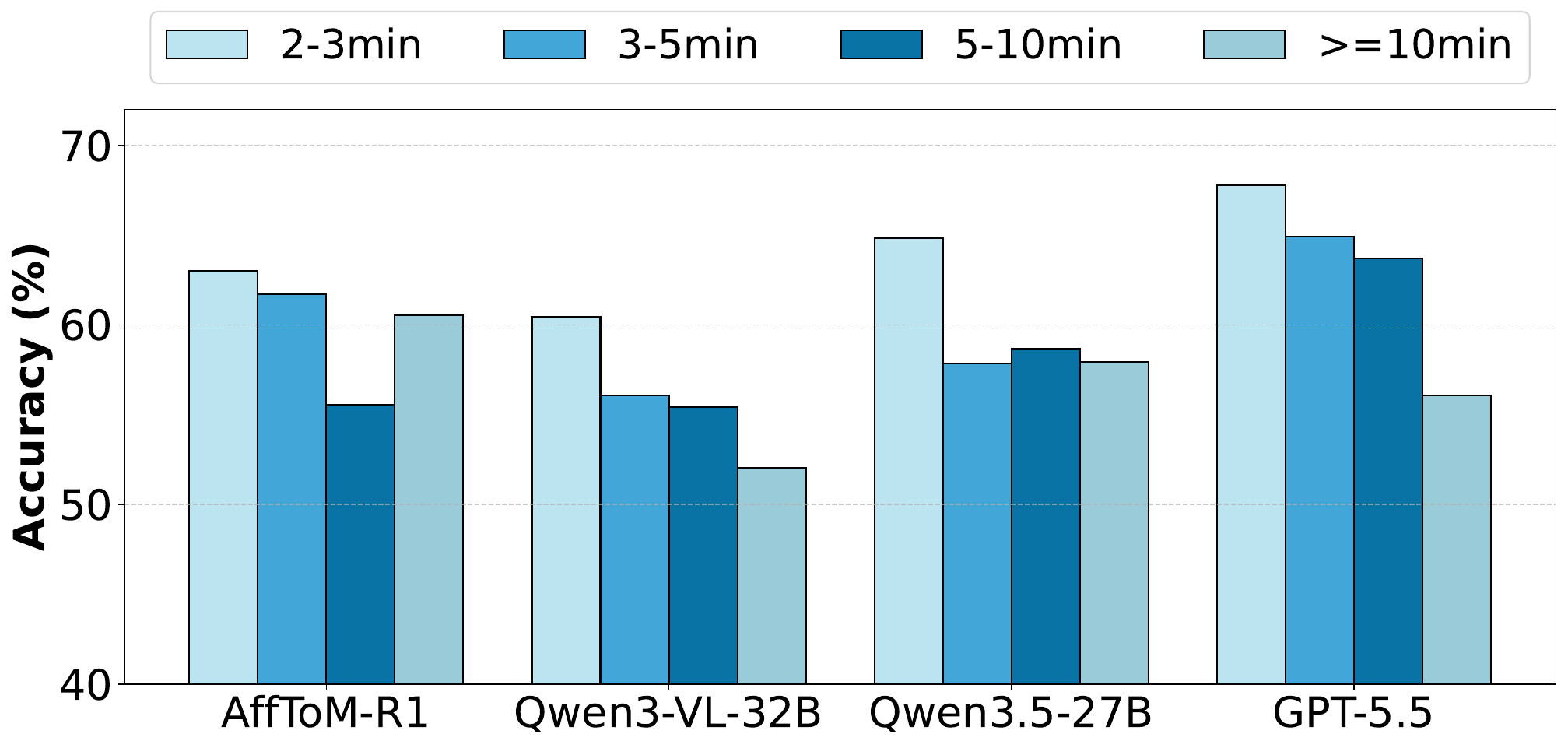}
    \caption{Accuracy across video durations.}
    \label{fig:video_duration_accuracy}
\end{figure}

\subsection{Effect of Frame Density and Video Duration}

\noindent  \textbf{Impact of Frame Density.}
We evaluate the impact of frame density by varying the number of uniformly sampled video frames from 0 to 64, where 0 frames corresponds to question-only evaluation.
The results are shown in Figure~\ref{fig:frame_count_accuracy}.
First, transitioning from 0 to 8 frames yields a sharp performance increase across all models, showing that even sparse visual cues are essential for reliably inferring narrative trajectories.
This upward trend generally continues up to 64 frames, further confirming the importance and effectiveness of denser spatiotemporal information.
Second, model capacity strictly dictates the ability to process these dense inputs.
While larger models consistently improve with more frames, smaller models plateau and exhibit slight fluctuations after 16 frames.
This saturation indicates a representational bottleneck in smaller networks, whereas larger models possess the capacity required to effectively integrate and reason over extended temporal sequences.

\noindent  \textbf{Impact of Video Duration.} We also investigate the effect of video duration, with results shown in Figure~\ref{fig:video_duration_accuracy}.
Longer videos are generally more challenging, especially for GPT-5.5 and Qwen3-VL-32B, whose accuracy declines as the duration bucket increases.
However, \methodName{} and Qwen3.5-27B exhibit non-monotonic changes across duration buckets, with partial recovery in the longest-video group.
Among the evaluated models, GPT-5.5 achieves the best overall performance.

\subsection{Effect of Subtitle Assistance}
\label{sec:subtitle_results}
\begin{table}[!t]
    \centering
    \small
    \setlength{\tabcolsep}{4pt}
    \renewcommand{\arraystretch}{1.12}
    \caption{
    Effect of subtitle assistance on \benchmarkName.
    }
    \label{tab:subtitle_results}
    \resizebox{0.99\linewidth}{!}{
    \begin{tabular}{
        @{}
        l
        !{\vrule width 0.5pt}
        c
        !{\vrule width 0.5pt}
        ccc
        !{\vrule width 0.5pt}
        ccc
        @{}
    }
        \toprule
        \multirow{2}{*}{\textbf{Models}}
        & \multirow{2}{*}{\textbf{Modality}}
        & \multicolumn{3}{c!{\vrule width 0.5pt}}{\textbf{MCQ}}
        & \multicolumn{3}{c}{\textbf{Open-ended}} \\
        \cmidrule(lr){3-5}
        \cmidrule(lr){6-8}
        &
        & \textbf{IAR}
        & \textbf{MPAR}
        & \textbf{Avg.}
        & \textbf{EC}
        & \textbf{FC}
        & \textbf{Avg.} \\
        \midrule

        \multirow{2}{*}{Qwen3.5-4B}
        & V
        & 57.83
        & 37.22
        & 49.29
        & 1.77
        & 1.50
        & 1.64 \\
        & V+T
        & \textbf{63.07}
        & \textbf{40.40}
        & \textbf{53.68}
        & \textbf{1.78}
        & \textbf{1.57}
        & \textbf{1.68} \\

        \midrule
        
        \multirow{2}{*}{\methodName{}}
        & V
        & 65.51 & 50.60 & 59.33
        & \textbf{2.17} & 1.89 & 2.03 \\
        & V+T
        & \textbf{66.35} & \textbf{52.72} & \textbf{60.70}
        & \textbf{2.17} & \textbf{1.96} & \textbf{2.07} \\

        \midrule

        \multirow{2}{*}{Qwen3.5-27B}
        & V
        & 69.26
        & 45.03
        & 59.22
        & 2.45
        & 2.23
        & 2.34 \\
        & V+T
        & \textbf{73.85}
        & \textbf{51.79}
        & \textbf{64.71}
        & \textbf{2.69}
        & \textbf{2.66}
        & \textbf{2.68} \\

        \midrule

        \multirow{2}{*}{GPT-5.5}
        & V
        & 71.70
        & 52.05
        & 63.56
        & 3.06
        & 2.99
        & 3.03 \\
        & V+T
        & \textbf{76.10}
        & \textbf{56.95}
        & \textbf{68.17}
        & \textbf{3.51}
        & \textbf{3.46}
        & \textbf{3.49} \\

        \bottomrule
    \end{tabular}
    }
\end{table}
We augment video inputs with subtitles generated by Whisper \textit{large-v2}~\citep{radford2022robustspeechrecognitionlargescale}.
Table~\ref{tab:subtitle_results} compares video-only inputs (V) with video-plus-subtitle inputs (V+T) across the evaluated models.
Subtitle assistance improves MCQ accuracy for every reported model, with the largest gains generally appearing on MPAR.
Its effects on open-ended EC and FC scores are mixed across models, with improvements observed for some larger baselines.
For the Qwen3.5-4B base model and \methodName{}, adding subtitles improves MCQ accuracy by 4.39 and 1.37 percentage points, respectively.
These results indicate that dialogue information provides complementary evidence for interpreting characters' beliefs and emotions.

\subsection{Analysis}
\label{sec_event_occur}

\noindent\textbf{Performance across Benchmarks.}
We evaluate the Qwen3.5-4B base model, its SFT variant, and \methodName{} on \benchmarkName{}, Social-IQ 2.0~\citep{siq2}, Hi-ToM~\citep{he2023hitombenchmarkevaluatinghigherorder}, and the DX and Hard subsets of R$^3$-Bench~\citep{niu2026read}.
These evaluations cover video-based social reasoning and text-based higher-order belief reasoning.
Table~\ref{tab:training_benchmarks} reports higher accuracy for both trained variants across all five evaluation sets, providing initial evidence that the benefits extend beyond \benchmarkName.
These gains suggest that learned reasoning transfers beyond the training distribution.
\begin{table}[t]
    \centering
    \caption{Accuracy (\%) across five benchmarks.
    Green values indicate gains over the base model in percentage points.}
    \label{tab:training_benchmarks}
    \scriptsize
    \setlength{\tabcolsep}{3pt}
    \renewcommand{\arraystretch}{1.2}
    \resizebox{\linewidth}{!}{
    \setlength{\aboverulesep}{0pt}
    \setlength{\belowrulesep}{0pt}
    \begin{tabular}{
        @{}
        l
        !{\vrule width 0.5pt}
        ccc
        @{}
    }
        \toprule
        \textbf{Dataset} & \textbf{Qwen3.5-4B} & \textbf{+SFT} & \textbf{\methodName{}} \\
        \midrule
        \benchmarkName
        & 49.29
        & 56.20\,\scalebox{0.85}{\textcolor{green!50!black}{(+6.91)}}
        & \textbf{59.33}\,\scalebox{0.85}{\textcolor{green!50!black}{(+10.04)}} \\

        Social-IQ 2.0
        & 65.13
        & 67.37\,\scalebox{0.85}{\textcolor{green!50!black}{(+2.24)}}
        & \textbf{69.99}\,\scalebox{0.85}{\textcolor{green!50!black}{(+4.86)}} \\

        Hi-ToM
        & 43.75
        & 49.33\,\scalebox{0.85}{\textcolor{green!50!black}{(+5.58)}}
        & \textbf{50.25}\,\scalebox{0.85}{\textcolor{green!50!black}{(+6.50)}} \\

        R$^3$-Bench-DX
        & 70.99
        & 75.55\,\scalebox{0.85}{\textcolor{green!50!black}{(+4.56)}}
        & \textbf{75.74}\,\scalebox{0.85}{\textcolor{green!50!black}{(+4.75)}} \\

        R$^3$-Bench-Hard
        & 32.98
        & 37.94\,\scalebox{0.85}{\textcolor{green!50!black}{(+4.96)}}
        & \textbf{39.72}\,\scalebox{0.85}{\textcolor{green!50!black}{(+6.74)}} \\
        \bottomrule
    \end{tabular}
    }
\end{table}

\noindent\textbf{Ablation Study.}
\label{sec:ablation_study}
We examine the contribution of these three reasoning rewards in \grpoName{} by enabling $R_{\mathrm{persp}}$, $R_{\mathrm{faith}}$, and $R_{\mathrm{cons}}$ while keeping the correctness and format rewards fixed.
As shown in Table~\ref{tab:reward_ablation}, $R_{\mathrm{persp}}$ provides the structural signal for reasoning from the required character viewpoint.
Its effect is most visible on MPAR, where perspective control separates a character's mental state from the video's full information.
Adding either $R_{\mathrm{faith}}$ or $R_{\mathrm{cons}}$ further improves the two-reward configuration, with faithfulness more closely supporting recursive mental-state tracking and consistency providing a stronger signal for direct affective inference.
Faithfulness grounds mental-state claims in the reference graph, while consistency links evidence, reasoning, and the answer.
The full reward set achieves the best results across IAR, MPAR, and Overall, reaching 65.51\%, 50.60\%, and 59.33\%, respectively.
These results support jointly enforcing perspective control, mental-state grounding, and evidence-answer consistency in \grpoName{}.

\begin{table}[t]
    \centering
    \makeatletter
    \def\@captype{table}
    \makeatother
    \caption{Ablation of PAFRO reward components on \benchmarkName.}
    \label{tab:reward_ablation}
    \scriptsize
    \setlength{\tabcolsep}{4pt}
    \renewcommand{\arraystretch}{1.2}
    \setlength{\aboverulesep}{0pt}
    \setlength{\belowrulesep}{0pt}
    \resizebox{\linewidth}{!}{
    \begin{tabular}{ccc!{\vrule width 0.5pt}ccc}
        \toprule
        \multicolumn{3}{c!{\vrule width 0.5pt}}{\textbf{Reward Components}}
        & \multicolumn{3}{c}{\textbf{Accuracy (\%)}} \\
        \midrule
        $R_{\mathrm{persp}}$ & $R_{\mathrm{faith}}$ & $R_{\mathrm{cons}}$
        & IAR & MPAR & Overall \\
        \midrule
        -- & -- & -- 
        & 64.88 & 47.35 & 57.61 \\
        
        \ding{51} & -- & -- 
        & 64.92 & 48.08 & 57.94 \\
        
        \ding{51} & \ding{51} & -- 
        & 65.04 & 49.80 & 58.73 \\
        
        \ding{51} & -- & \ding{51} 
        & 65.32 & 49.01 & 58.56 \\
        
        \ding{51} & \ding{51} & \ding{51}
        & \textbf{65.51} & \textbf{50.60} & \textbf{59.33} \\
        \bottomrule
    \end{tabular}}
\end{table}

\begin{figure}[t]
    \centering
    \includegraphics[width=\linewidth]{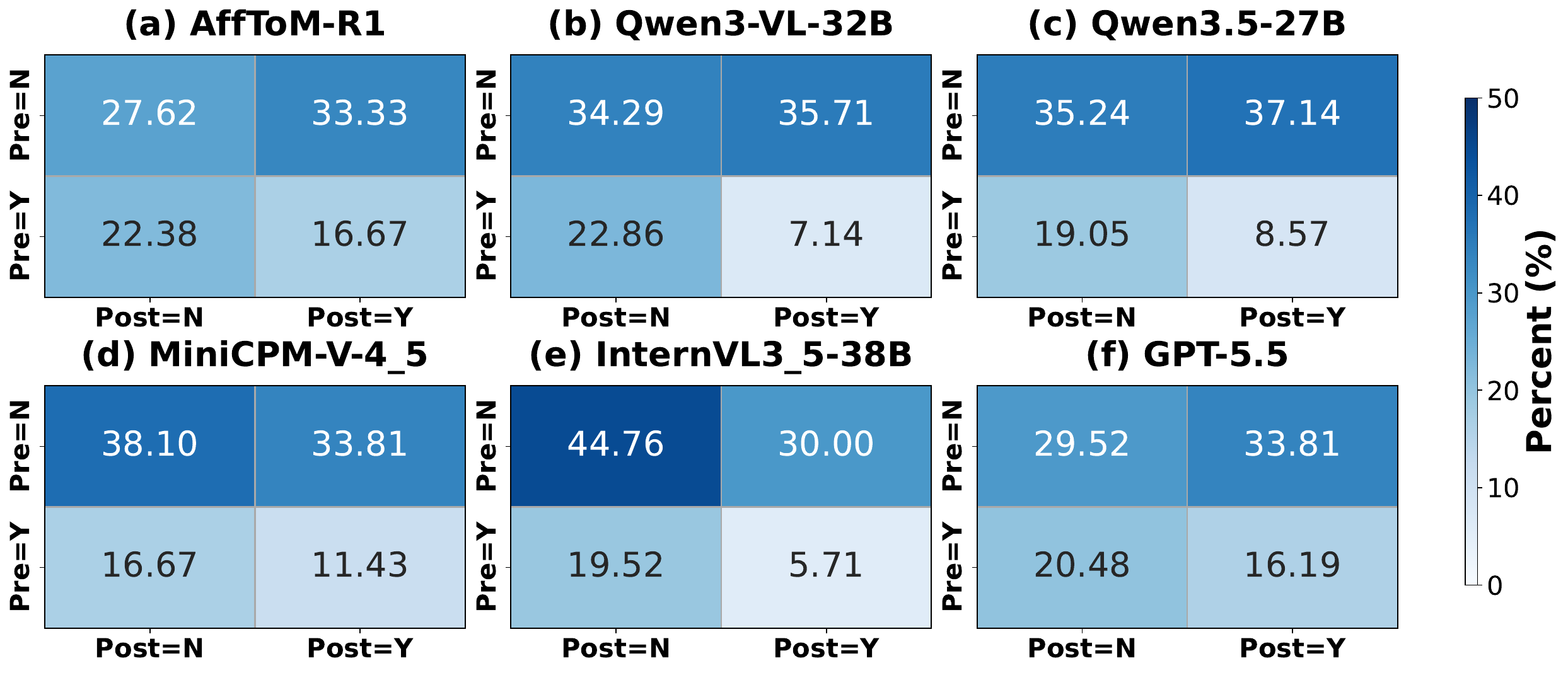}
    \caption{Joint correctness on paired MPAR questions. Rows show pre-event results and columns show post-event results. Y denotes a correct answer, and N an incorrect one.}
    \label{fig:temporal_compare}
\end{figure}

\noindent  \textbf{Effect of Information Revelation.}
Information asymmetry plays a critical role in shaping emotional dynamics.
We conduct an analysis comparing model performance on pre-event and post-event questions, with results shown in Figure ~\ref{fig:temporal_compare}.
Pre-event questions require models to simulate a restricted cognitive state and reason under incomplete information, where characters hold divergent beliefs or unresolved misunderstandings.
In contrast, post-event questions evaluate affective states after the interaction unfolds and the epistemic uncertainty is largely resolved.
We observe that models consistently achieve better performance on post-event questions.
The higher proportion of Pre-N/Post-Y than Pre-Y/Post-N cases suggests that models are more likely to reason from an omniscient perspective, using later revealed information rather than preserving the character's pre-event knowledge state.
These findings highlight a limitation of current models in handling information asymmetry, which is essential for robust affective ToM understanding.

\begin{figure}[t]
    \centering
    \includegraphics[width=\linewidth]{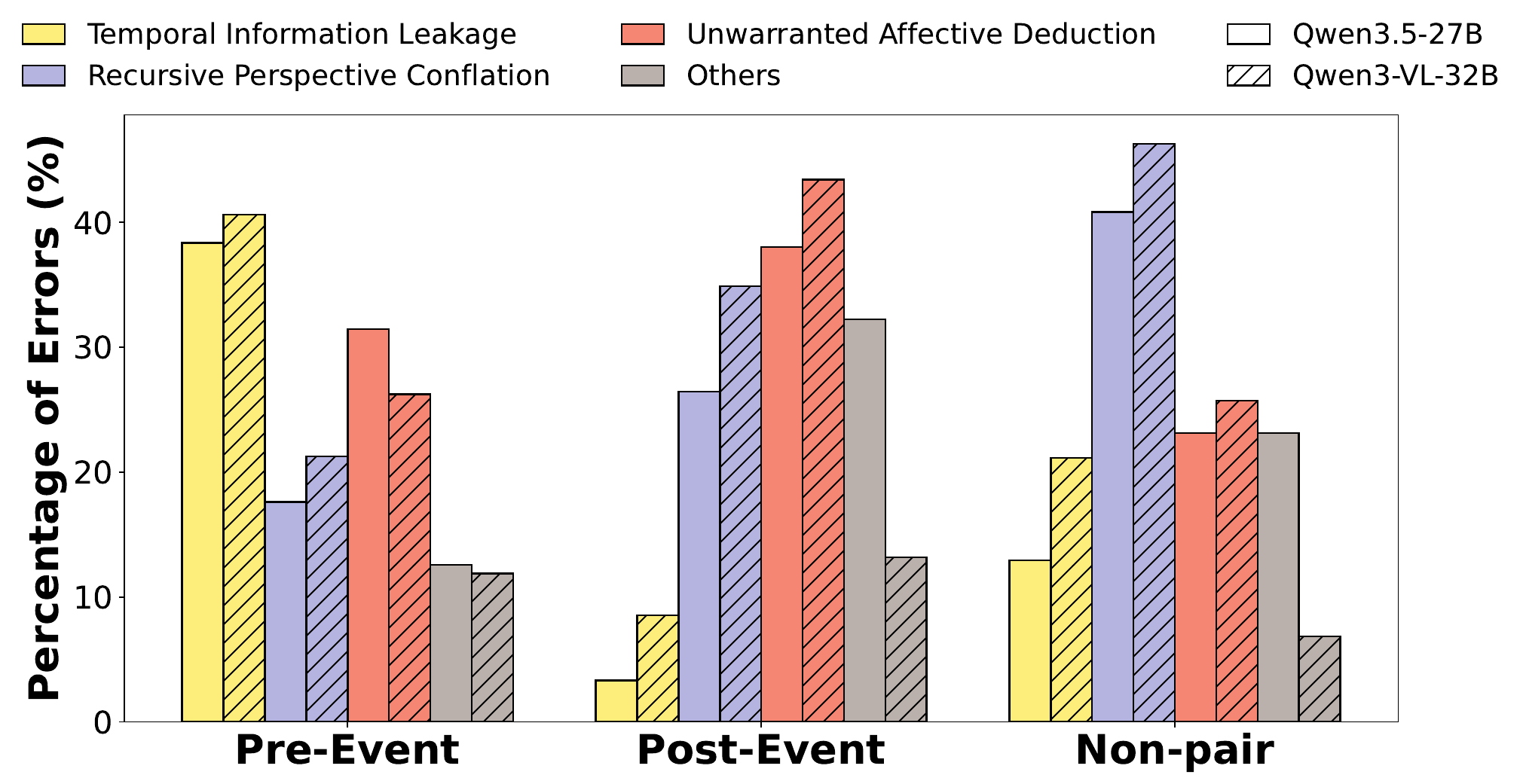}
    \caption{MPAR MCQ error distribution (\%). Non-Pair denotes questions outside the pre/post-event pairs.}
    \label{fig:error_analysis}
\end{figure}

\noindent\textbf{Error Analysis.}
To further characterize model failures, we analyze the generated reasoning outputs of Qwen3.5-27B and Qwen3-VL-32B on their respective incorrectly answered MPAR MCQs and classify the resulting failures into four categories: \textit{{\OmniscientLeakage}}, \textit{{\Perspective}}, \textit{{\Faulty}}, and \textit{{\Others}}, as summarized in Figure~\ref{fig:error_analysis}.
For \textbf{Pre-Event} questions, \textit{{\OmniscientLeakage}} is the most prominent error, indicating that models tend to rely on information revealed at later event stages when reasoning under asymmetric information.
For \textbf{Post-Event} questions, \textit{{\Faulty}} becomes the dominant error type.
This shift demonstrates that while models can observe the physical event, they struggle to accurately map these visual observations to updated psychological states, frequently hallucinating unsupported affective dependencies or over-interpreting sparse cues.
For \textbf{Non-pair} questions, \textit{{\Perspective}} emerges as the most severe bottleneck.
Absent explicit temporal anchoring, models persistently conflate objective multimodal facts with the target character's subjective, nested beliefs.
Notably, despite differences in overall model capacity, Qwen3.5-27B and Qwen3-VL-32B exhibit highly consistent error distributions across all three temporal conditions.
Overall, these findings indicate that higher-order affective reasoning is mainly limited by temporal grounding under asymmetric information, faithful recursive perspective tracking, and psychologically plausible inference.
Representative failure cases are discussed in Appendix~\ref{app:case_study}.

\section{Conclusion}
In this paper, we propose {\benchmarkName}, a pioneering multimodal benchmark specifically designed to evaluate affective ToM capabilities in dynamic multi-party social interactions.
Moving beyond traditional epistemic reasoning and isolated emotion recognition, our dataset challenges models to decipher latent affective states within dynamic, multi-party social interactions.
We construct the benchmark through an event-driven multi-agent generation pipeline grounded in dynamic character interaction graphs, followed by dual-criteria evolution and human validation to ensure data quality.
To further investigate whether MPAR can be improved through targeted post-training, we introduce \methodName{}, which combines SFT with perspective-aware rubric-guided optimization.
Results across diverse MLLM families show that MPAR remains challenging, while \methodName{} delivers a substantial 13.38-percentage-point MPAR gain over its Qwen3.5-4B base model and consistent gains on external ToM and social-reasoning benchmarks.

\bibliography{custom}

\appendix
\section{Dataset Details}
\label{app:dataset_details}

\subsection{Dataset Statistics}
\label{app:dataset_statistics}

\noindent  \textbf{Video Sources and Duration.}
{\benchmarkName} contains 258 videos collected from public video-sharing platforms, including 186 videos from Douyin/TikTok and 72 videos from YouTube.
The dataset has a total duration of approximately 26.4 hours, with an average video length of 6.1 minutes.
The collected videos primarily consist of social narratives, including short dramas and movie/TV drama clips, which naturally contain rich interpersonal interactions and evolving affective dynamics.
The dataset includes videos in both Chinese and English, reflecting a range of social interaction contexts.

\begin{figure}[t]
    \centering
    \includegraphics[width=\linewidth]{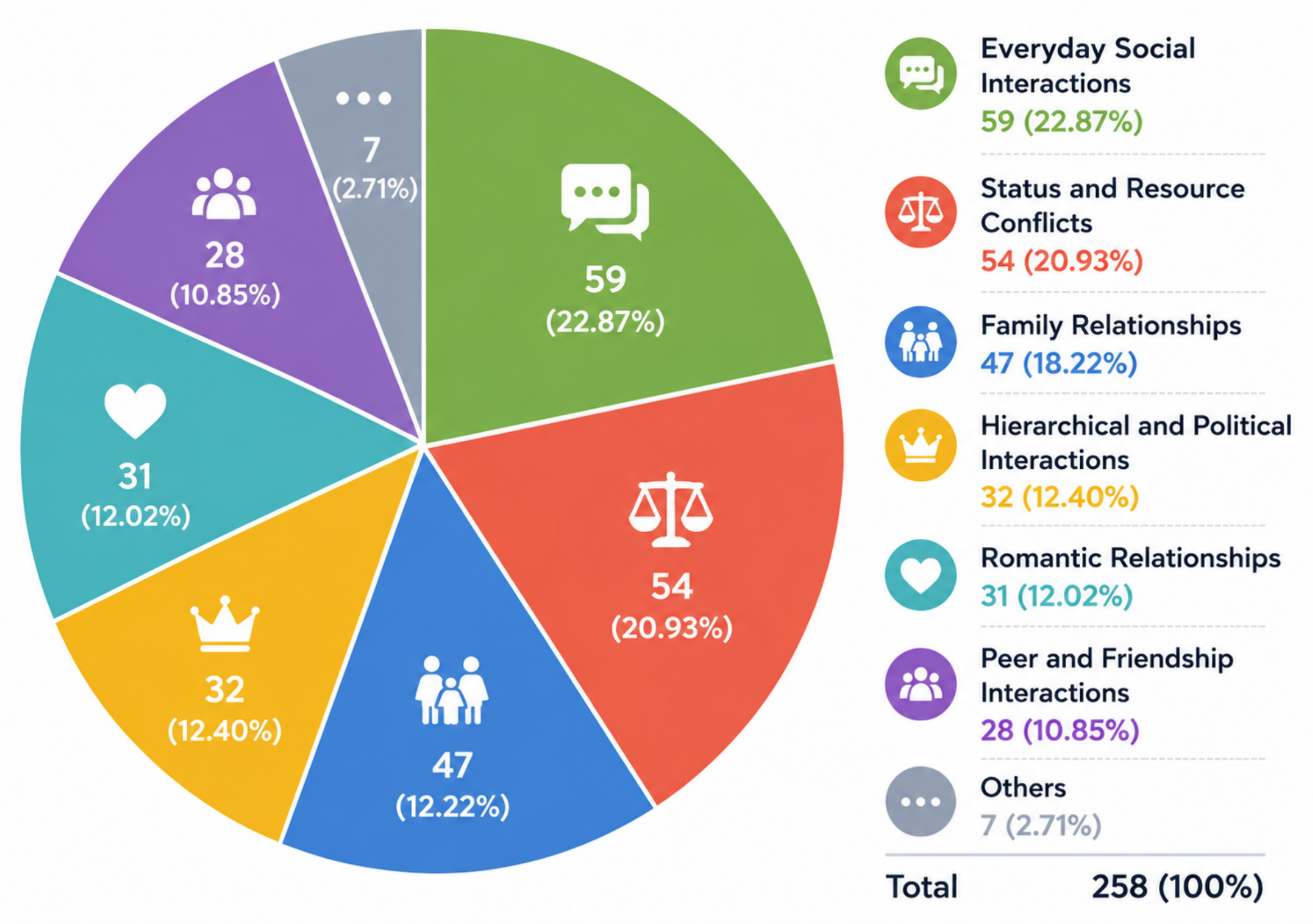}
    \caption{Scenario distribution of {\benchmarkName}.}
    \label{fig:data_details}
\end{figure}

\noindent  \textbf{Scenario Diversity.} As illustrated in Figure~\ref{fig:data_details}, {\benchmarkName} covers a wide range of social scenarios rather than concentrating on a single interaction pattern.
The collected videos span diverse interpersonal contexts, including family relationships, status/resource conflicts, hierarchical and political interactions, romantic relationships, peer friendships, and everyday social interactions.
Such diversity provides rich social dynamics for evaluating affective Theory of Mind under realistic multi-party interactions.

\noindent \textbf{Filtering Statistics.}
During dataset construction, the multi-agent pipeline initially produced 4,174 question-answer pairs from 308 filtered videos.
After automatic filtering for text-only shortcuts and option-level biases, 2,340 candidates were retained for human validation.
At this stage, all QA candidates associated with 50 videos were filtered out.
Consequently, these videos were excluded, reducing the dataset from 308 to 258 videos.
Human annotators subsequently spent approximately 150 hours reviewing and validating the retained candidates.

\begin{figure*}[!htbp]
    \centering
    \includegraphics[width=1.0\linewidth]{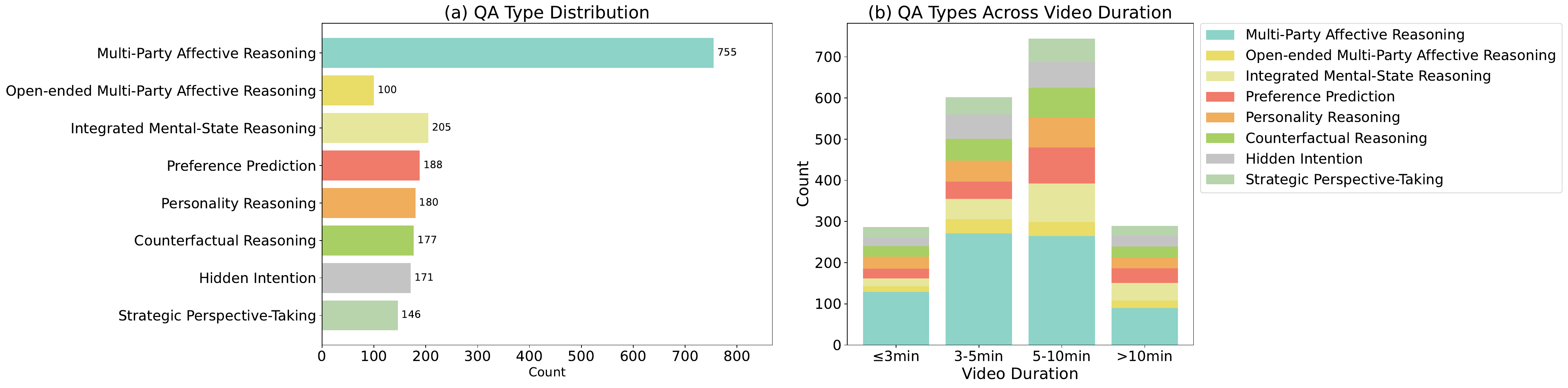}

    \caption{Distribution of QA types in the dataset and the distribution of QA types across different video-duration ranges.}
    \label{fig:qa_distribution}
\end{figure*}

\noindent  \textbf{Question Taxonomy and Distribution.}
After filtering and human validation, {\benchmarkName} contains 1,922 question-answer pairs, including 1,822 multiple-choice questions and 100 open-ended questions.
The questions are organized into two broad categories: Individual Affective Reasoning (IAR) and Multi-party Affective Reasoning (MPAR).
IAR questions are presented in a multiple-choice format and cover six task paradigms: Preference Prediction, Integrated Mental-State Reasoning, Strategic Perspective-Taking, Hidden Intention, Personality Reasoning, and Counterfactual Reasoning.
MPAR includes both multiple-choice and open-ended questions that evaluate recursive perspective tracking and belief-conditioned affective inference.
Figure~\ref{fig:qa_distribution} presents the overall distribution of question types, together with their distribution across different video-duration ranges.

\subsection{Training Data Sources}
\label{app:training_data_sources}

The auxiliary training pool comprises 2,291 videos: 735 self-collected videos and 1,556 CondensedMovies clips from the $R^3$-FDT video collection~\citep{niu2026read}.
Applying our multi-agent data construction pipeline to these videos yields 19,759 QA samples, including 6,490 from the self-collected videos and 13,269 from CondensedMovies.
We select subsets of this pool for SFT and PAFRO, with training settings detailed in Appendix~\ref{app:training_settings}.

\section{Human Validation and Evaluation}
\label{app:human_evaluation}

\subsection{Dataset Validation}
\label{app:human_validation}

Four students with backgrounds in computer science manually review the generated questions.
Before large-scale annotation, they independently score a preliminary subset on four dimensions: answer reasonableness, video dependency, distractor quality, and question validity.
Each dimension uses a 5-point Likert scale, ranging from 1 (poor quality) to 5 (superior quality).
Substantial disagreements are discussed under the guidance of domain experts to establish a consistent evaluation standard.
After calibration, all annotators independently score {\evaluationNum} randomly sampled instances to assess inter-annotator agreement.
Table~\ref{tab_human_evaluation} reports ICC(3,k) scores above 0.80 across all four dimensions, indicating strong agreement.
The annotators then verify the remaining question-answer pairs against the raw videos using the same four criteria, discarding any instance with an average score below 3.5.

\begin{table}[t]
\centering
\small
\setlength{\tabcolsep}{7pt}
\caption{Inter-annotator agreement measured by ICC(3,k) across evaluation dimensions. ANSR, VDEP, DQLT, and QVAL denote answer reasonableness, video dependency, distractor quality, and question validity, respectively.}
\label{tab_human_evaluation}
\resizebox{0.99\linewidth}{!}{
\begin{tabular}{@{}lccccc@{}}
\toprule
\textbf{Metric} & \textbf{ANSR} & \textbf{VDEP} & \textbf{DQLT} & \textbf{QVAL} & \textbf{AVG.} \\
\midrule
\textbf{ICC (3,k)} & 0.811 & 0.823 & 0.852 & 0.810 & \textbf{0.824} \\
\bottomrule
\end{tabular}
}
\vspace{-1em}
\end{table}

\subsection{Human Baseline Evaluation}
\label{app:human_baseline}

To evaluate human performance on \benchmarkName, we recruited five graduate students with backgrounds in computer science and multimodal learning as human testers.
Although affective states may admit semantically related interpretations, each evaluated multiple-choice question was manually validated to contain a single best-supported answer
grounded in the video context.
Before the formal evaluation, all testers independently answered the same 30 randomly sampled questions without time constraints.
Across this shared calibration subset, no pair of testers differed on more than three questions, indicating broadly consistent task interpretation.
For the formal evaluation, we randomly sampled 250 multiple-choice questions and partitioned them evenly into five non-overlapping subsets, with each tester answering one subset.
Thus, each question received one human response.
Aggregating the responses across all subsets yielded 222 correct answers out of 250 questions, corresponding to an overall accuracy of 88.80\%.

\subsection{LLM-as-a-Judge Validation}
\label{app:judge_validation}

To validate GPT-5-mini as the automatic evaluator for open-ended responses, a human annotator independently scores 100 responses generated by Qwen3.5-27B and another 100 responses generated by Qwen3-VL-32B using the same EC and FC rubrics.
For Qwen3.5-27B, the Spearman correlations between human and GPT-5-mini scores are 0.766 for EC and 0.727 for FC.
For Qwen3-VL-32B, they are 0.790 for EC and 0.769 for FC.
These correlations support the use of GPT-5-mini for automatic scoring under the EC and FC rubrics.

\section{Experimental Implementation Details}
\label{app:implementation_details}

\subsection{Training Settings}
\label{app:training_settings}

We implement \methodName{} in ms-swift\footnote{\url{https://github.com/modelscope/ms-swift}} using Qwen3.5-4B as the backbone.
Both stages use AdamW, a constant learning rate schedule without warmup, bfloat16 precision, and a random seed of 42.
Training inputs contain 16 video frames and the available subtitle text.
Question difficulty is estimated from the pretrained backbone's accuracy over eight rollouts per question.

\noindent\textbf{SFT optimization.}
Using the retained SFT subset, we train for two epochs with a learning rate of $2\times10^{-5}$ and a batch size of 16.

\noindent\textbf{PAFRO.}
The PAFRO policy and its fixed reference are initialized from the SFT-trained model.
We train on 5k QA pairs for one epoch with a learning rate of $2\times10^{-6}$ and a KL coefficient of $0.04$.
We use a batch size of 32 and generate four rollouts per question at temperature $1.0$.
Qwen3.5-27B serves as the reward model for the faithfulness and consistency rubrics.
The reward weights $(\lambda_{\mathrm{acc}},\lambda_{\mathrm{fmt}},\lambda_{\mathrm{persp}},\lambda_{\mathrm{faith}},\lambda_{\mathrm{cons}})$ are set to $(1.0,0.1,0.1,0.4,0.2)$.

\subsection{SFT Trajectory Construction}
\label{app:sft_construction}
We construct each SFT trajectory with two successive calls to Qwen3.5-27B.
In the first call, a perspective-classification prompt provides the video, question, and answer options, and asks the model to determine whether solving the question requires focused reasoning about the target character's information state and mental state.
Questions that require character-specific beliefs or mental-state attribution are assigned the \emph{character-centered} label, whereas questions that can be answered from the broader event context are assigned the \emph{context-centered} label.
In the second call, we provide the same video-question input together with the predicted perspective label and ask the model to generate a reasoning trajectory in the corresponding format.
For \emph{character-centered} questions, the trajectory follows \emph{Clues} $\rightarrow$ \emph{Mental State} $\rightarrow$ \emph{Inference}.
For \emph{context-centered} questions, it follows \emph{Clues} $\rightarrow$ \emph{Inference}.
The second prompt also constrains the response to include a \texttt{<perspective>}\allowbreak\texttt{</perspective>} field, a reasoning block enclosed by \texttt{<think>}\allowbreak\texttt{</think>}, and a final answer enclosed by \texttt{<answer>}\allowbreak\texttt{</answer>}.
All MPAR questions use the \emph{character-centered} structure, while the remaining questions receive the structure selected in the first step.
We retain trajectories whose final answers match the reference answers, and split the resulting approximately 20k-sample pool into SFT and PAFRO subsets according to the difficulty estimated from the pretrained backbone.

\subsection{PAFRO Optimization with GRPO}
\label{app:grpo_optimization}

For each video-question pair $(v,q)\in\mathcal{D}_{\mathrm{RL}}$, the rollout policy $\pi_{\mathrm{old}}$ samples a group of $K$ responses $\{y_i\}_{i=1}^{K}$.
Each response receives the weighted reward $r_i$ in Equation~\ref{eq:combined_reward}, which combines rule-based checks with the two rubric scores.
We compare responses within the same group rather than fitting a separate value model.
Writing $\bar r=K^{-1}\sum_{j=1}^{K}r_j$, the response-level advantage is
\begin{equation}
    \hat A_i=
    \frac{r_i-\bar r}
    {\operatorname{Std}(r_1,\ldots,r_K)+\delta},
    \label{eq:grpo_advantage}
\end{equation}
where $\delta>0$ stabilizes groups with little reward variation.
Thus, a response is favored when its reward exceeds that of other responses to the same input.

The trainable policy $\pi_\theta$ and fixed reference policy $\pi_{\mathrm{ref}}$ are both initialized from the SFT-trained model, while $\pi_{\mathrm{old}}$ denotes the policy used to generate the current rollout group.
For token $y_{i,t}$, let
$\rho_{i,t}=\pi_\theta(y_{i,t}\mid v,q,y_{i,<t})/
\pi_{\mathrm{old}}(y_{i,t}\mid v,q,y_{i,<t})$
be the probability ratio between the updated and rollout policies.
The clipped token objective is
\begin{equation}
    \resizebox{0.85\linewidth}{!}{$\displaystyle
        \ell_{i,t}(\theta)=
        \min\!\left(
            \rho_{i,t}\hat A_i,\,
            \operatorname{clip}(\rho_{i,t},1-\epsilon,1+\epsilon)\hat A_i
        \right),
    $}
\end{equation}
where $\epsilon$ limits the size of policy changes.
We maximize
\begin{equation}
    \resizebox{0.85\linewidth}{!}{$\displaystyle
        \mathcal{J}_{\mathrm{GRPO}}(\theta)
        =\mathbb{E}\!\left[
            \frac{1}{K}\sum_{i=1}^{K}\frac{1}{|y_i|}
            \sum_{t=1}^{|y_i|}
            \left(\ell_{i,t}(\theta)-\beta d_{i,t}\right)
        \right],
    $}
    \label{eq:grpo_objective}
\end{equation}
where $d_{i,t}$ is the sampled token-level KL penalty relative to $\pi_{\mathrm{ref}}$, and $\beta$ sets its weight.
The expectation covers training inputs and their sampled response groups.
Responses receive equal weight, and tokens are averaged within each response.
The same response-level advantage applies to all tokens of a rollout, linking the reward for its answer and reasoning to the policy update.
Clipping and the fixed-reference penalty constrain the update while the group comparison favors higher-reward trajectories.

\subsection{Reward Definitions}
\label{app:reward_definitions}

This section details the five reward components combined in Equation~\ref{eq:combined_reward}.

\noindent\textbf{Correctness reward $R_{\mathrm{acc}}$.}
We compare the extracted answer $a(y)$ with the reference $a^{\ast}$:
\begin{equation}
    R_{\mathrm{acc}}(y)=\mathbb{I}\!\left(a(y)=a^{\ast}\right),
    \label{eq:reward_accuracy}
\end{equation}
where $\mathbb{I}(\cdot)$ is the indicator function defined above.
Invalid answer extractions receive zero reward.

\noindent\textbf{Format reward $R_{\mathrm{fmt}}$.}
This reward preserves the response structure learned during SFT.
Let $\mathcal{F}$ denote responses that match the prescribed structure over the entire output.
The format is:
\begin{quote}
\ttfamily
<think>\\
<perspective>\normalfont\textit{label}\ttfamily</perspective>\\
\normalfont\textit{reasoning}\ttfamily\\
</think>\\
<answer>\normalfont\textit{answer letter}\ttfamily</answer>
\end{quote}
The \texttt{<think>} block must begin with the perspective field followed by reasoning, and the subsequent \texttt{<answer>} block must contain a single answer letter.
Whitespace variations are permitted, but additional text outside these blocks is not allowed.
This reward checks structural compliance.
Perspective alignment and answer correctness are evaluated separately.
We define
\begin{equation}
    R_{\mathrm{fmt}}(y)=\mathbb{I}\!\left(y\in\mathcal{F}\right).
    \label{eq:reward_format}
\end{equation}

\noindent\textbf{Perspective reward $R_{\mathrm{persp}}$.}
This reward compares the generated perspective label with the annotated route and target character.
We extract the predicted label from the \texttt{<perspective>} field and normalize superficial differences in case and formatting.
\emph{Context-centered} questions require a \emph{context-centered} prediction, while \emph{character-centered} questions require a character-specific label sharing at least one character-descriptive token with the normalized reference.
This lexical matching accommodates variation in character names and descriptions without requiring exact string equality.
The reward is:
\begin{equation}
    R_{\mathrm{persp}}(y)=
    \begin{cases}
        1, & \text{matched},\\
        0, & \text{otherwise}.
    \end{cases}
    \label{eq:reward_perspective}
\end{equation}
Missing or ambiguous perspective extractions receive zero reward.

\noindent\textbf{Faithfulness reward $R_{\mathrm{faith}}$.}
The reward model assesses the accuracy and relevance of mental-state claims by comparing them with the dynamic character interaction graph $G$ provided as reference, using the sample's sub-event or whole-video scope.
This reward applies to both routes, assessing \emph{context-centered} reasoning without imposing a single-character viewpoint.
For \emph{character-centered} responses, it evaluates these claims from the viewpoint named in the generated label.
We normalize the reward model's assessment under the faithfulness rubric to obtain
\begin{equation}
    R_{\mathrm{faith}}(y)=\operatorname{Norm}\!\left(\mathcal{M}_{\mathrm{faith}}(q_{\mathrm{text}},p(y),\tau_{\mathrm{r}}(y),\mathcal{G})\right),
    \label{eq:reward_faithfulness}
\end{equation}
where $\operatorname{Norm}(z)=z/2$ maps the raw rubric score $z\in\{0,1,2\}$ to $[0,1]$.

\noindent\textbf{Consistency reward $R_{\mathrm{cons}}$.}
The multimodal reward model evaluates whether the reasoning in $y$ supports its final answer using evidence from video $v$ and question $q$.
It receives sampled frames and associated subtitles as scoring evidence, without the reference answer.
The reward is
\begin{equation}
    R_{\mathrm{cons}}(y)=\operatorname{Norm}\!\left(\mathcal{M}_{\mathrm{cons}}(v,q,y)\right),
    \label{eq:reward_consistency}
\end{equation}
using the same normalization, with $q$ including the answer options and $y$ containing both the reasoning and the final answer.
The rubric favors clear evidential support and penalizes contradictions or unsupported conclusions.
The functions $\mathcal{M}_{\mathrm{faith}}$ and $\mathcal{M}_{\mathrm{cons}}$ denote separate calls to the reward model, each using its own rubric prompt to produce a response-level score.

\section{Case Study}
\label{app:case_study}
\begin{figure*}[!htbp]
    \centering
    \includegraphics[width=1.0\linewidth]{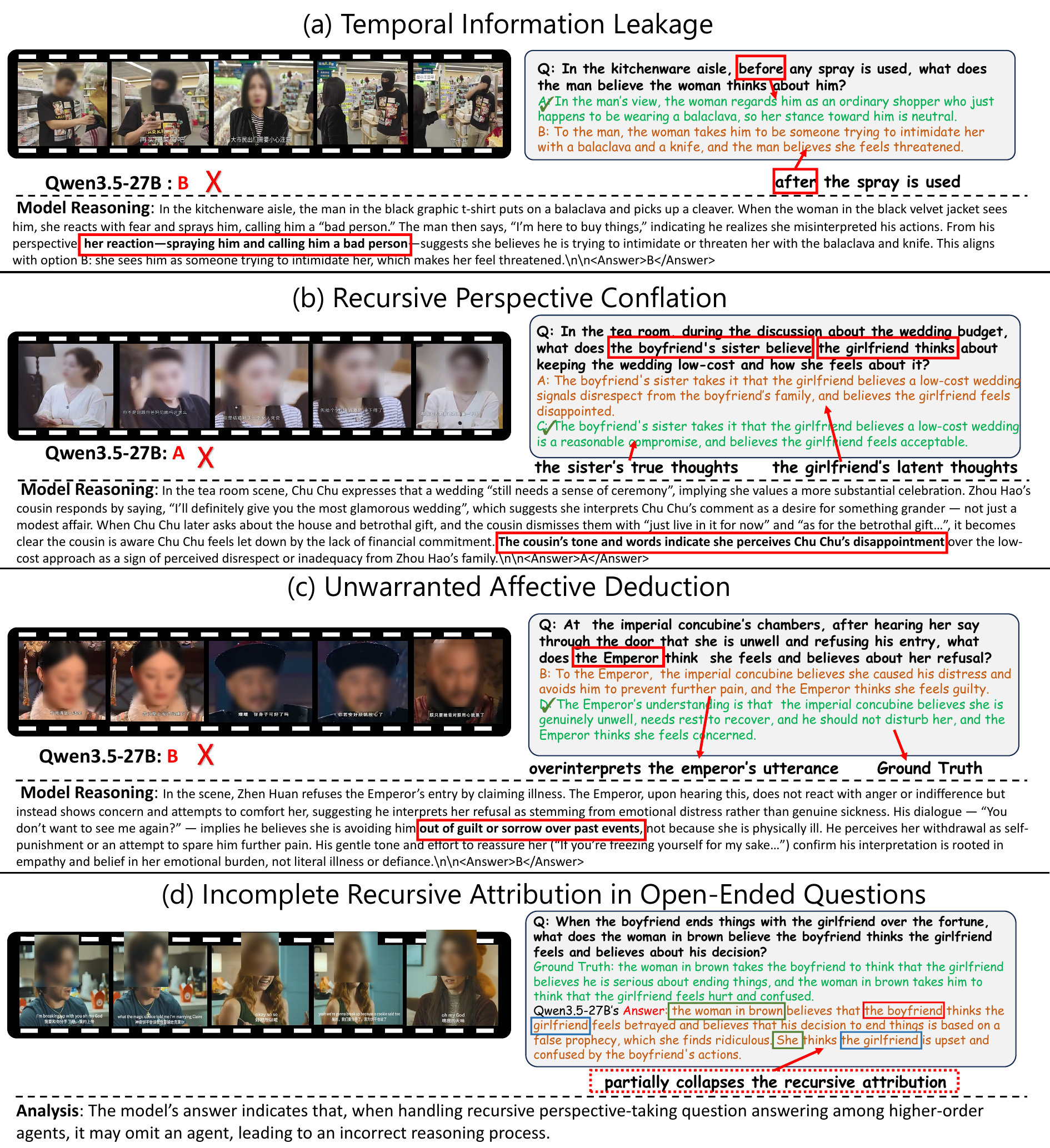}
    \caption{Representative failure cases of MLLMs on {\benchmarkName}, including three major error types in MPAR—Temporal Information Leakage, Recursive Perspective Conflation, and Unwarranted Affective Deduction—as well as a typical error in open-ended question answering involving missing recursive characters.}
    \label{fig:case_study}
\end{figure*}

Figure~\ref{fig:case_study} presents representative failures that illustrate the error patterns above and the degradation on open-ended questions.
In Figure~\ref{fig:case_study} (a), the model's reasoning trace reveals a clear temporal information leakage error.
Although the question is anchored before any spray is used, the model justifies its answer by referring to the woman's later reaction, namely that she sprays the man and calls him a bad person.
This evidence is unavailable at the queried moment: before the spray, the man has not yet inferred that the woman mistakes him for a robber or feels threatened by him.
By using this post-event reaction to infer the man's pre-event belief, the model answers from an omniscient timeline rather than from the man's temporally constrained perspective.

In Figure~\ref{fig:case_study} (b), the model's reasoning trace reveals a recursive perspective conflation error.
The question targets the boyfriend's sister's belief about the girlfriend's thoughts and feelings, but the model grounds its answer in the girlfriend's own latent dissatisfaction.
It cites Chu Chu's remarks about needing a ceremonial wedding and her concerns about the house and betrothal gift, and then infers that the boyfriend's sister's tone and words show she perceives Chu Chu's disappointment.
In this way, the model treats the girlfriend's internal disappointment as if it were already represented in the sister's belief.
The predicted answer therefore conflates the girlfriend's true inner state with the sister's second-order perspective, rather than reasoning only from what the sister can observe and infer.

In Figure~\ref{fig:case_study} (c), the model's reasoning trace exposes an unwarranted affective deduction rather than a temporal or perspective-tracking error.
The model stays within the Emperor's current viewpoint, but overinterprets his dialogue and tone.
From his concern and the utterance ``You don't want to see me again?'', it infers that he believes the concubine is avoiding him out of guilt, sorrow, self-punishment, or a desire to spare him pain.
This goes beyond the available evidence: the scene only supports that she claims illness and that the Emperor responds with concern for her condition.
The failure is therefore an unsupported evidence-to-emotion mapping, where sparse cues are turned into a strong but unjustified emotional-causal chain.

In Figure~\ref{fig:case_study}(d), the model initially preserves the outer belief chain but partially collapses the recursive attribution in its subsequent explanation.
Specifically, the final statement directly attributes the girlfriend's emotional state to the woman in brown, rather than maintaining the woman's belief about the boyfriend's inference.
The response therefore represents the recursive structure only incompletely.

\section{Multi-Agent Data Construction}
\label{app:system_details}

\subsection{System Configuration}
\label{app:system_configuration}

Our multi-agent system is implemented using the LangGraph framework
\footnote{\url{https://github.com/langchain-ai/langgraph}} and consists of six agents: the Event Agent, State Agent, Question Generation Agent, Role Reflection Agent, Revision Agent, and Evaluation Agent.
The Event Agent and State Agent extract structured events and character-level mental states, which are integrated into $G$ for subsequent question generation and refinement.
Based on preliminary experiments, we use Seed-1.8 as the backbone of the Event Agent and State Agent due to its stronger semantic extraction capability, while the remaining reasoning and refinement agents use Qwen3-VL-32B as their backbone.

All agents use $\textit{top\_p}=0.95$.
The Event Agent, State Agent, and Evaluation Agent use
$\textit{temperature}=0$ for deterministic extraction and evaluation,
while the remaining agents use $\textit{temperature}=0.7$ for diverse
generation, reflection, and revision.
During iterative refinement, we randomly sample
$|\mathcal{Q}_{\text{human}}|=5$ human-authored reference questions
from the seed pool for pairwise comparison.
We set the acceptance threshold to $\tau=0.8$, the trade-off coefficient
to $\lambda=0.7$, and the maximum number of refinement rounds to
$T_{\max}=3$.
For recursive perspective expansion, we set the maximum recursive depth to $d_{\max}=3$, balancing reasoning complexity and annotation reliability.

\subsection{MPAR Question Generation Example} 
\label{app:mpar_generation_example}
\subsubsection{Event Extraction Output}
Here we present an example of an event list extracted from a single
video.
It records the main characters, event timestamps, locations, event
summaries, and involved individuals.
The event list directly provides the event-level context for base-level question generation and, together with the corresponding individual-level information, is used to construct the dynamic character interaction graph $G$ for recursive perspective expansion.

Event List Example:
\begin{tcolorbox}[breakable]
\begin{lstlisting}[breaklines=true, basicstyle=\ttfamily\small]
"character_list": [{
  "character_id": "Wang Zha",
  "description": "A young man in a black jacket, trying to reconcile with his girlfriend Leng Meng and later proposing to her. His parents misinterpret his interactions with his friend."
},
{
  "character_id": "Leng Meng",
  "description": "Wang Zha's girlfriend, wearing a white coat, sitting upset on a park bench after a misunderstanding with Wang Zha."
},
{
  "character_id": "Leng Meng's male friend",
  "description": "A man in a denim jacket with glasses, protecting Leng Meng from Wang Zha initially, and later mistakenly targeted by Wang Zha's parents."
},
{
  "character_id": "Wang Zha's father",
  "description": "An older man in a gray jacket, who misinterprets Wang Zha's interactions with his male friend as a romantic relationship and becomes distraught."
},
{
  "character_id": "Wang Zha's mother",
  "description": "An older woman in a beige coat, who tries to calm her husband and also misinterprets Wang Zha's interactions, later offering money to Leng Meng's friend."
}],
"events": [
{
  "event_id": "event_1",
  "timestamp": "00:00 - 00:37",
  "location": "Outdoor park area",
  "description": "Wang Zha tries to approach the upset Leng Meng, but her male friend blocks him. Wang Zha argues with the friend, using casual terms like 'dear' which are overheard by Wang Zha's parents. The parents misinterpret the interaction as Wang Zha being in a romantic relationship with the male friend.",
  "interaction_pairs": [
    ["Wang Zha", "Leng Meng's male friend"],
    ["Leng Meng's male friend", "Wang Zha"],
    ["Leng Meng's male friend", "Leng Meng"],
    ["Wang Zha", "Leng Meng"],
    ["Leng Meng", "Wang Zha"],
    ["Wang Zha's father", "Wang Zha's mother"],
    ["Wang Zha's mother", "Wang Zha's father"],
    ["Wang Zha's father", "Wang Zha"],
    ["Wang Zha's mother", "Wang Zha"]
  ]
},
...]

\end{lstlisting}
\end{tcolorbox}

\subsubsection{Graph Initialization and Recursive Question Generation}

\begin{figure}[!htbp]
    \centering
    \includegraphics[width=0.99\linewidth]{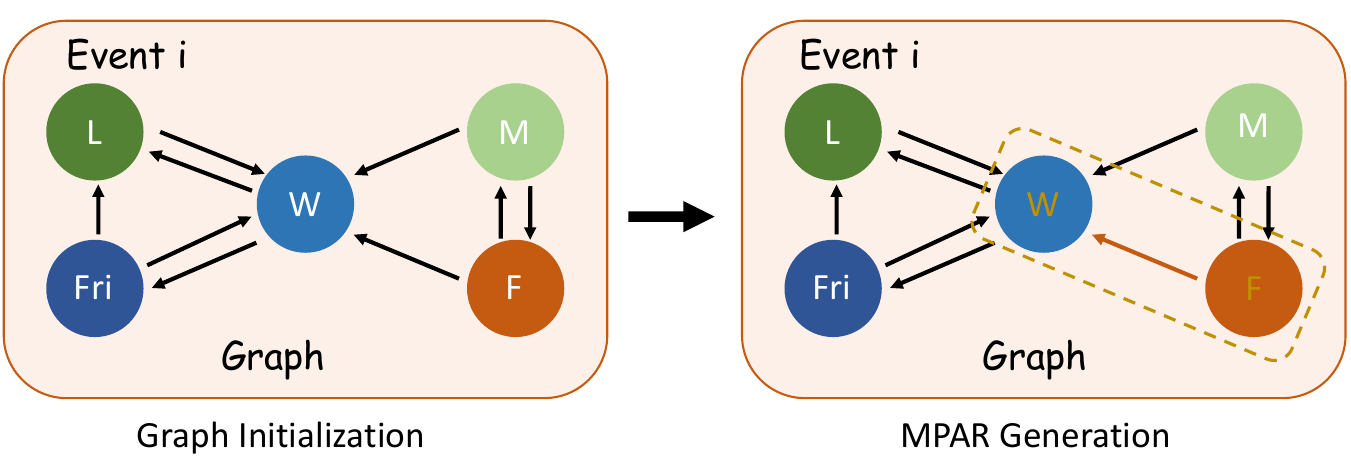}
    \caption{An example of a character interaction graph. The nodes represent ``L'': ``Leng Meng'', ``Fri'': ``Leng Meng's male friend'', ``W'': ``Wang Zha'', ``M'': ``Wang Zha's mother'', and ``F'': ``Wang Zha's father''. The left panel shows the initialized graph, while the right panel highlights the characters selected for recursive perspective construction during MPAR generation.}
    \label{fig:node}
\end{figure}

Based on the extracted events $\mathcal{E}$ and individual-level information $\mathcal{S}$, we construct the dynamic character interaction graph $G$, as illustrated in Figure~\ref{fig:node}.
Each node represents an involved character, while each directed edge indicates the direction of an observed interaction.
The left panel of Figure~\ref{fig:node} shows the initialized graph, and the right panel illustrates how it supports Recursive Perspective Expansion for MPAR question generation.

We first establish a base-level affective state or belief for a target character grounded in the event context.
In the running example, Wang Zha is selected as the target character.
To expand the reasoning structure, we identify characters who directly interact with or observe the current character within the current event context encoded in $G_{i}$ and treat them as candidate outer observers.
When multiple candidates are available, the LLM jointly examines the graph $G_{i}$, the event-localized video segment, and the current recursive belief chain.
The selection considers whether the event supports information asymmetry between the candidate and the current subject, whether the candidate's belief concerns the current subject, and whether the associated emotion, intention, or desire is relevant to the target relation.
The selected character is then introduced as the new outer observer, extending the recursive perspective chain.

In Figure~\ref{fig:node}, Wang Zha's father, represented by node \(F\), is selected as the outer observer.
This expands the reasoning structure from Wang Zha's base-level mental state to a second-order perspective.
The graph $G_{i}$, the corresponding video segment, and the current belief chain are subsequently provided to the Question Generation Agent, which produces an MPAR question-answer pair together with plausible distractors.

\subsection{Pipeline Robustness Audit}
\label{app:pipeline_audit}

To further assess the robustness of the multi-agent construction pipeline, we conduct an additional audit of representative intermediate outputs.
During dataset construction, intermediate supervision is incorporated into the pipeline: before large-scale generation, we inspect stage-level outputs and proceed only after the outputs become stable.
To quantitatively examine whether upstream errors could systematically affect final QA quality, we further re-audit 20 randomly sampled generation traces across three stages that are most relevant to final QA validity: event extraction, role/mental-state extraction, and recursive QA construction.

We apply stage-specific criteria to inspect intermediate outputs at the level of individual checkable positions.
Each occurrence that violates a criterion contributes one error count, and the error rate is computed as the total number of errors divided by the total number of corresponding positions inspected across the 20 generation traces.
For event extraction, the inspected positions include character-interaction edges, event temporal boundaries, character identities, and required output format.
Errors are counted for incorrectly constructed interaction edges, temporal intervals that substantially deviate from the event’s reasonable boundaries, misidentified characters, or violations of the prescribed output format.
For role/mental-state extraction, we inspect character profiles, event-specific mental-state entries for each present character, and required output format.
Errors include inaccurate profile information, missing mental-state entries for characters involved in an event, mental-state analyses that are clearly inconsistent with the video evidence, and formatting violations.
For recursive QA construction, we inspect every belief-extraction step along the recursive perspective chain, together with the required output format.
Each incorrect belief extraction or formatting violation contributes one error count.

As shown in Table~\ref{tab:pipeline_audit}, most inspected intermediate positions are consistent with the video, although residual errors remain across all three audited stages.
Role/mental-state extraction shows an error rate of 5.22\%, reflecting occasional inaccurate mental-state analysis in sub-events.
Recursive QA construction has an error rate of 7.41\%, mainly caused by unsupported relational or affective links.
Given the limited sample size and the position-level nature of this audit, these results should be interpreted as a diagnostic analysis rather than a dataset-wide reliability estimate.
The observed errors, particularly unsupported relational or affective links, are further screened during final human validation, where annotators verify, revise, or discard QA candidates based on the original video evidence and event-localization information.

\begin{table}[!htbp]
    \centering
    \small
    \caption{Pipeline robustness audit on 20 randomly sampled generation traces. Error rates are computed over all inspected intermediate positions.}
    \label{tab:pipeline_audit}
    \resizebox{0.99\linewidth}{!}{
    \begin{tabular}{lll}
        \toprule
        Stage & Primary Observed Error & Error Rate \\
        \midrule
        Event Extraction 
        & Incorrect interaction edges 
        & 3.67\% \\
        Role/Mental-State Extraction 
        & Incorrect mental-state analysis in events 
        & 5.22\% \\
        Recursive QA Construction 
        & Unsupported relational/affective links 
        & 7.41\% \\
        \bottomrule
    \end{tabular}
    }
    \vspace{-1em}
\end{table}

\section{Prompt Details}
\label{app:prompt_details}

\subsection{Prompt for Multiple-Choice Questions}

\begin{tcolorbox}[breakable]
\textbf{\# Instruction}:\\
These are the frames of a video.

\vspace{0.5em}

(Optional) The video's subtitles are provided below when subtitle assistance is enabled:

\vspace{0.5em}

\{Subtitles\}

\vspace{0.5em}

Select the best answer to the following multiple-choice question. Respond with the exact final sentence ``\texttt{<answer>X</answer>}'', where X is one of A, B, C or D.

\vspace{0.5em}

\textbf{\# Question}: \{Question\}

\vspace{0.5em}

\textbf{\# Options}:\\
\textbf{A}: \{Option A\}\\
\textbf{B}: \{Option B\}\\
\textbf{C}: \{Option C\}\\
\textbf{D}: \{Option D\}

\vspace{0.5em}

\textbf{\# Answer}:
\end{tcolorbox}

\subsection{Prompt for Open-Ended Questions}

\begin{tcolorbox}[breakable]
\textbf{\# Instruction}:\\
These are the frames of a video.

\vspace{0.5em}

(Optional) The video's subtitles are provided below when subtitle assistance is enabled:

\{Subtitles\}

Please answer the following question.

\vspace{0.5em}

\textbf{\# Question}: \{Question\}

\vspace{0.5em}

\textbf{\# Task}:\\
Your answer must explicitly describe a three-level belief structure. Specifically, you should express Agent A’s interpretation of Agent B’s belief about Agent C’s perspective on the event, as well as Agent A’s inference of Agent B’s belief about Agent C’s emotional attitude toward the event.
\end{tcolorbox}

\subsection{Prompt for LLM-as-a-Judge in Open-Ended questions}
\begin{tcolorbox}[breakable]
\textbf{\# Instruction}\\
You are a helpful assistant in evaluating open-ended Q\&As. Please act as an unbiased judge. 
The reference answer has been manually validated against the source video and serves as the factual anchor for evaluation. Evaluate the assistant's response based only on the question and the human-validated reference answer. 
Assess whether the response preserves the factual content, character roles, and recursive viewpoint structure represented in the reference answer. Do not introduce or reward additional interpretations solely because they appear plausible.

\vspace{0.5em}

\textbf{\# Notes}\\
- Your evaluation should determine whether the assistant effectively follows the requirements of the question and correctly addresses it.\\
- You must assess the response across the two dimensions and score each dimension independently.\\
- Do not let the length of the response influence your judgment.\\
- Do not favor any specific assistant identity or position. Remain as objective as possible.\\
- Do not penalize wording differences when the response is semantically equivalent to the reference answer.

\vspace{0.5em}

\textbf{\# Criteria}\\
Use scores to represent the quality of the assistant's response. Below are the detailed scoring rubrics.

\vspace{0.5em}

\textbf{\#\# Emotion Correctness (EC)}\\
This dimension evaluates whether the response correctly identifies the emotional state of the target character, consistent with the reference answer. Factual alignment of the emotion is the primary criterion.\\
- \textbf{1 -- Poor:} The predicted emotion is entirely opposite to the reference answer. Even if the nested perspective is grammatically fluent, an opposite emotion must receive the lowest score.\\
- \textbf{2 -- Fair:} The predicted emotion shows weak relevance and contains noticeable errors, such as identifying an incorrect or overly weak emotion.\\
- \textbf{3 -- Moderate:} The general emotional direction is somewhat correct, but lacks clarity, specificity, or accurate recursive attribution.\\
- \textbf{4 -- Good:} The emotional polarity is highly consistent with the reference answer, with only minor differences in wording or intensity.\\
- \textbf{5 -- Excellent:} The predicted emotion fully matches the reference answer in polarity, intensity, and semantic precision, and is correctly attributed within the recursive belief structure.

\vspace{0.5em}

\textbf{\#\# Factual Consistency (FC)}\\
This dimension evaluates the accuracy of the embedded viewpoint content and whether the response correctly represents the recursive belief structure (i.e., Agent A's perspective on Agent B's belief about Agent C's view).\\
- \textbf{1 -- Poor:} The core viewpoint contradicts the reference answer or the recursive chain is missing or incorrect.\\
- \textbf{2 -- Fair:} The viewpoint is largely inaccurate, focused on the wrong event, or severely misinterprets the characters' intentions.\\
- \textbf{3 -- Moderate:} The recursive structure is identifiable, but the viewpoint only partially aligns with the reference and contains omissions or minor role confusion.\\
- \textbf{4 -- Good:} The recursive chain is clear and correctly ordered. The viewpoint closely matches the reference answer, with only minor semantic differences.\\
- \textbf{5 -- Excellent:} The recursive structure is explicit, correctly ordered, and unambiguous. The embedded viewpoint perfectly matches the reference answer in meaning, focus, and interpretation.

\vspace{0.5em}

\textbf{\# Test Question}\\
\textbf{QUESTION:} \{question\_text\}\\
\textbf{ASSISTANT ANSWER:} \{pred\_answer\}\\
\textbf{GROUND TRUTH:} \{ground\_truth\}

\vspace{0.5em}

\textbf{\# Output Format}\\
You MUST output your evaluation strictly in the following JSON format:
\begin{lstlisting}[breaklines=true, basicstyle=\ttfamily\small]
{
    "Emotion_Correctness": <score from 1 to 5>,
    "Factual_Consistency": <score from 1 to 5>
}
\end{lstlisting}

\end{tcolorbox}

\end{document}